\documentclass[10pt,twocolumn,letterpaper]{article}

\usepackage[pagenumbers]{cvpr} 

\usepackage{graphicx}
\usepackage{amsmath}
\usepackage{amssymb}
\usepackage{booktabs}
\usepackage{kotex}
\usepackage{times}
\usepackage{epsfig}
\usepackage{mathtools}
\usepackage{dsfont}
\usepackage{cuted} 
\usepackage{tabularx}
\usepackage{comment}
\usepackage[accsupp]{axessibility}
\usepackage{multirow}
\usepackage{subcaption}
\usepackage{float}
  \floatstyle{ruled}
  \newfloat{algorithm}{thp}{alg}
\floatname{algorithm}{Algorithm}
\usepackage{algorithmicx}
\usepackage{bm}
\usepackage{caption}
\usepackage[dvipsnames]{xcolor}
\definecolor{SigmaColor}{rgb}{0.98,0.45,0.0}

\newcommand\sbullet[1][.5]{\mathbin{\vcenter{\hbox{\scalebox{#1}{$\bullet$}}}}}

\usepackage[pagebackref,breaklinks,colorlinks]{hyperref}

\usepackage[capitalize]{cleveref}
\crefname{section}{Sec.}{Secs.}
\Crefname{section}{Section}{Sections}
\Crefname{table}{Table}{Tables}
\crefname{table}{Tab.}{Tabs.}

\newcommand\blfootnote[1]{
  \begingroup
  \renewcommand\thefootnote{}\footnote{#1}
  \addtocounter{footnote}{-1}
  \endgroup
}

\def\cvprPaperID{3509} 
\def\confName{CVPR}
\def\confYear{2024}

\begin{document}

\title{Mocap Everyone Everywhere: \\
Lightweight Motion Capture With Smartwatches and a Head-Mounted Camera}

\author{Jiye Lee\\
Seoul National University\\
{\tt\small kay2353@snu.ac.kr}
\and
Hanbyul Joo\\
Seoul National University\\
{\tt\small hbjoo@snu.ac.kr}
} 

\maketitle
\begin{strip}\centering
\includegraphics[width=\linewidth, trim={0cm 0cm 0 0cm},clip]{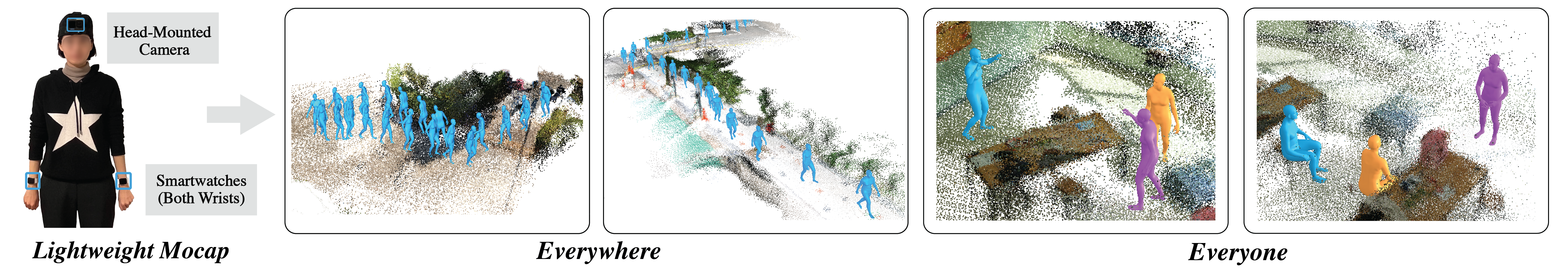}
\captionof{figure}{We present a lightweight and affordable motion capture method based on two smartwatches and a head-mounted camera.
}
\label{fig:teaser}
\end{strip}

\begin{abstract}
We present a lightweight and affordable motion capture method based on two smartwatches and a head-mounted camera. In contrast to the existing approaches that use six or more expert-level IMU devices, our approach is much more cost-effective and convenient. Our method can make \textbf{wearable motion capture accessible to everyone everywhere}, enabling 3D full-body motion capture in diverse environments.
As a key idea to overcome the extreme sparsity and ambiguities of sensor inputs with different modalities, we integrate 6D head poses obtained from the head-mounted cameras for motion estimation.
To enable capture in expansive indoor and outdoor scenes, we propose an algorithm to track and update floor level changes to define head poses, coupled with a multi-stage Transformer-based regression module.
We also introduce novel strategies leveraging visual cues of egocentric images to further enhance the motion capture quality while reducing ambiguities.
We demonstrate the performance of our method on various challenging scenarios, including complex outdoor environments and everyday motions including object interactions and social interactions among multiple individuals.
\blfootnote{Project Page: \href{https://jiyewise.github.io/projects/MocapEvery}{https://jiyewise.github.io/projects/MocapEvery}}
\end{abstract}
\vspace{-20pt}
\section{Introduction}
\label{sec:intro}

Human motion encapsulates the diverse and rich stories of human life in real-world situations. Thus, it is essential for machines not only to decipher the subtle nuances within these movements but also to authentically mimic them for seamless interaction and assistance with humans.
However, the advancement in the research to make machines fully understand human motions lags considerably behind other AI domains focused on images or languages. 
The primary obstacle lies in the fact that 3D human motion datasets that reflect real-world scenarios remain exceedingly scarce, compared to the abundance of massive image or language datasets that capture diverse aspects of human life.
This scarcity stems from the inherent challenge of acquiring motion capture data, necessitating specialized equipment and tools that are not readily available to the public. Despite notable efforts in collecting and refining motion capture datasets that are actively used in learning-based approaches (e.g., \cite{cmumocap, mahmood2019amass}), the scale of these datasets remains in orders of magnitude smaller than the vast repositories of Internet images and language datasets. 

In our endeavor to \emph{democratize} motion capture for the general public free from the constraints of expert-level equipment and intricate settings, we present a lightweight and affordable motion capture method that utilizes Inertial Measurement Unit (IMU) sensors of smartwatches on wrists and a head-mounted camera. 
Our method can make wearable motion capture accessible to everyone, given the widespread availability and popularity of smartwatches, as well as the existence of action cameras and camera glasses (e.g., GoPro or Smart Glasses~\cite{metarayban, metaaria} equipped with cameras). Our system can be applied for 
 large-scale and long-term motion captures without location constraints (indoors or outdoors). To this end, our method can enable the exploration of new research areas, such as social interaction scenarios involving multiple participants, each equipped with our lightweight capture setup.

Going beyond the traditional optical motion capture methods~\cite{vicon, joo2015panoptic}, primarily feasible in well-set lab environments with limited scope and accessibility, there have been explorations to capture human motions with wearable sensors. 
Existing commercial solutions~\cite{xsens, rokoko} require attaching IMU sensors to all major body segments (17 to 19) using specialized suits or straps. While recent start-of-the-art approaches~\cite{yi2021transpose, yi2022pip, jiang2022tip} have reduced the necessity to 6 sensors, it is still challenging for the public to equip themselves with 6 expert-level IMU sensors. Moreover, attaching sensors to the pelvis and legs may compromise usability.
Due to the rise of VR/AR technologies and hardware, another research direction~\cite{du2023agrol, jiang2022avatarposer, winkler2022questsim, lee2023questenvsim, aliakbarian2023hmdnemo} utilize head-mounted displays (HMDs)\cite{metaquest} and hand-held controllers, which provide 6DoF (3D rotation and translation) information of the head and wrist movements, to reconstruct full-body motions. Using only three upper body sensors may improve usability, but this reliance on specialized devices restricts motion capture to indoor settings optimized for VR/AR equipment, thus limiting its broader applicability.

In this paper, we present a motion capture method that utilizes IMU sensors of smartwatches on wrists and a head-mounted monocular camera. We overcome the sparsity of using only 2 sensors and the intrinsic ambiguities of IMU sensors (orientation, acceleration) by integrating head 6DoF poses obtained from the head-mounted camera into the motion estimation pipeline. 
In contrast to VR settings (e.g., HMDs) where head poses are given in a fixed world coordinate in a small indoor environment~\cite{du2023agrol, jiang2022avatarposer, winkler2022questsim, lee2023questenvsim, aliakbarian2023hmdnemo}, it is not trivial to define head poses in expansive outdoor settings. 
Through an algorithm that automatically 
tracks and adjusts floor levels, 
our approach can robustly capture motion in ``in-the-wild" scenarios in challenging locations with non-flat grounds such as hilly areas or stairs.
Furthermore, we also present a motion optimization approach for enhancing capture quality in complex everyday motions, such as object interactions, by leveraging observed body part cues through the head-mounted camera.
As our system can be conveniently equipped by multiple users, we also explore scenarios where the signals among the individuals are shared and the egocentric observations from one person can be used as sparse third-person views for the others, which can be additionally used for our motion optimization module. 

Our contribution can be summarized as follows:
(1) the first method to capture high-quality 3D full-body motion from a head-mounted camera and two smartwatches on wrists. Notably, we demonstrate our method on commonly available smartwatches rather than expert-level IMU sensors;
(2) a novel algorithm to track and update floor levels coupled with a multi-stage Transformer-based motion estimation module, which enables capture in expansive indoor and outdoor settings;
(3) a novel motion optimization module that utilizes visual information captured by monocular egocentric cameras.

\section{Related Work}
\label{sec:relwork}

\noindent \textbf{Motion Capture with IMU Sensors.} 
Traditional motion capture mainly relied on optical methodologies such as optical markers~\cite{vicon} or markerless capture in a multi-view camera setup~\cite{liu2013markerless, joo2015panoptic, zhang2021lightweight}. Although such methods demonstrate high-quality capture, they suffer from occlusions and are only applicable in specific settings with cameras. To mitigate such limitations there have been methods to fuse inertial sensor data to optical motion capture~\cite{gilbert2019fusing, helten2013real,kaichi2020resolving,malleson2017real, malleson2020real,kalkbrenner2014imukinect,von2016human,pons2010multisensor,von20183dpw, trumble2017totalcapture}. 
Wearable motion capture with IMU sensors offers the advantage of freedom from location constraints and occlusions. Commercial methods~\cite{xsens, rokoko} leverage 17 to 19 IMU sensors, necessitating tight suits or straps with densely packed sensors.
More recent approaches introduce more lightweight solutions by reducing the number of sensors up to 6, posed on the pelvis, wrists, legs, and head~\cite{von2017sip, huang2018dip, yi2021transpose, yi2022pip, jiang2022tip, yi2023egolocate}. SIP~\cite{von2017sip} introduces an optimization-based approach and DIP~\cite{huang2018dip} presents a deep learning-based method from 6 sensors using bidirectional RNNs. TransPose~\cite{yi2021transpose} extends this work by including global root translation estimation based on foot contact detection. More recently, PIP~\cite{yi2022pip} and TIP~\cite{jiang2022tip} demonstrate state-of-the-art performance by combining physics-based optimization with an RNN-based kinematic approach~\cite{yi2022pip} or combining body contact estimation and terrain generation with Transformer decoders~\cite{jiang2022tip}.
To address the inherent root drift issue in IMU-based methods, there have been approaches that additionally utilize camera positions from head-mounted cameras to enhance the accuracy of root translations~\cite{guzov2021hps, yi2023egolocate}. In these methods, the camera poses are often used as auxiliary cues to post-process root translation, while human pose estimation is still based on full-body IMU sensor setup (17 for~\cite{guzov2021hps} or 6 for~\cite{yi2023egolocate}). Typically, previous methods are demonstrated using expert-level devices~\cite{xsens}. Different from previous approaches, our method uses two smartwatches and integrates a head-mounted camera into the body pose estimation pipeline.

\noindent \textbf{Motion Capture with Egocentric Videos.}
Pose estimation from egocentric videos is receiving increased attention in recent studies. 
Several methods exploit fish-eye cameras by leveraging their advantages in visibility for egocentric-pose estimation~\cite{jiang2021egocentric, tome2020selfpose, tome2019xr, wang2021estimating, xu2019mo, wang2023scene}. Cha et al.~\cite{cha2021mobile} extend these methods by fusing full-body IMU sensors with fish-eye camera input. 
A few approaches pursue more challenging scenarios by estimating human pose with body-mounted cameras without other sensors.
Shiratori et al.~\cite{shiratori2011mocap} use structure-from-motion (SfM) to determine body poses from body-mounted cameras, and Jiang et al.~\cite{jiang2017seeing} exploit motion graphs for estimation from a single chest-mounted camera. Other methods~\cite{yuan20183d, yuan2019ego} combine kinematics and physics-based approaches to estimate physically plausible poses, which is extended by~\cite{luo2021kinpoly, merel2020catch} to simple scene interactions. You2Me~\cite{ng2020you2me} estimates pose under social interaction scenarios where the interaction target's pose is visible in the egocentric video. Apart from these methods that directly learn from image inputs, EgoEgo~\cite{li2023egoego} estimates motions from an egocentric video alone by introducing head poses as intermediate representations.

\noindent \textbf{Motion Capture with VR-Based Upper Body Sensors.}
Recent advances in VR/AR technologies have led to the emergence of new devices that provide 6DoF information on head and wrist movements. Consequently, research has been done on controlling virtual avatars through such devices~\cite{yang2021lobstr, dittadi2021fullbody, ahuja2021coolmoves, aliakbarian2022flag, jiang2022avatarposer, du2023agrol, aliakbarian2023hmdnemo, winkler2022questsim, lee2023questenvsim}. As determining full body motion from upper body sensors is challenging, earlier methods~\cite{yang2021lobstr, dittadi2021fullbody} use root information explicitly by adding sensors on the pelvis~\cite{yang2021lobstr} or implicitly by root-relative data representations~\cite{dittadi2021fullbody}. CoolMoves~\cite{ahuja2021coolmoves} suggests kNN-based search which is limited to specific action types. More recent approaches demonstrate increased performance by exploiting various neural network structures such as flow models~\cite{aliakbarian2022flag}, Transformers~\cite{jiang2022avatarposer}, and diffusion models~\cite{du2023agrol}. QuestSim~\cite{winkler2022questsim} utilizes physics-based character control using deep reinforcement learning (DRL) to generate physically plausible full-body motions, which is extended by~\cite{lee2023questenvsim} to environment interactions.
It is important to note that motion estimation with VR devices fundamentally differs from ours, as the 6DoF cues from the head and wrists are directly provided by the system and thus rely on a specified setting.

\section{Method}
\label{sec:methods}

\begin{figure*}
\centering
\includegraphics[width=0.95\linewidth, trim={0 0.8cm 0 0}]{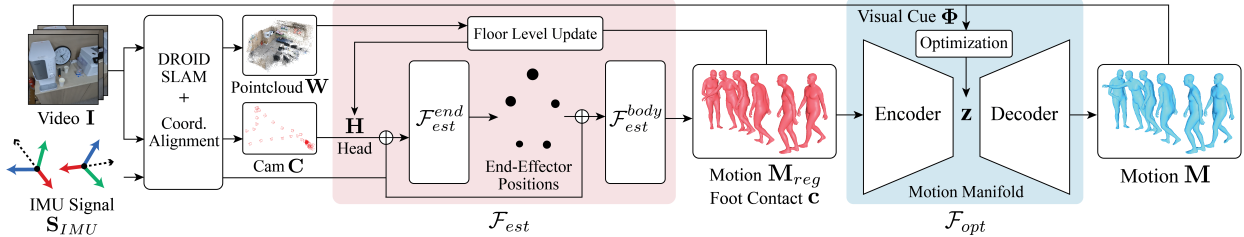}
    \caption{System Overview}
    \label{fig:sys_overview}
\vspace{-0.4cm}
\end{figure*}

\subsection{Overview}
Our method reconstructs 3D full-body human motion $\mathbf{M}=\{\mathbf{m}_t\}_{t=0}^T$ by taking, as inputs, an egocentric video $\mathbf{I}$ and IMU sensor signals $ \mathbf{S}_{IMU} = \{ \mathbf{R}, \mathbf{A} \}$ (orientation, acceleration) from two smartwatches on wrists. We denote our system as a function $\mathcal{F}$,
\begin{equation}
    \mathbf{M} = \mathcal{F}(\mathbf{I}, \mathbf{S}_{IMU})
\end{equation}
where $\mathcal{F}$ is composed of two modules $\mathcal{F}_{est}$ and $\mathcal{F}_{opt}$, as shown in Fig.~\ref{fig:sys_overview}.
The egocentric video from a head-mounted camera is a sequence of images, $\mathbf{I} = \{\mathbf{I}_t\}_{t=0}^{T}$ where $T$ is the sequence length, and $\mathbf{I}_t \in \mathbb{R}^{h \times w \times 3}$ indicates an image at time $t$.  
$\mathbf{S}_{IMU} = \{ \mathbf{R}, \mathbf{A} \}$ is a set of IMU signal sequences for both the left and right wrist. 
$\mathbf{R} = \{\mathbf{R}_t\}_{t=0}^T$, where 
$\mathbf{R}_t = (\mathbf{r}_t^{left}, \mathbf{r}_t^{right})$, indicates the orientation signal for each sensor at time $t$ in angle axis,  $\mathbf{r}_t \in so(3)$. 
Similarly, $\mathbf{A} = \{\mathbf{A}_t \}_{t=0}^T$, where $\mathbf{A}_t =(\mathbf{a}_t^{left}, \mathbf{a}_t^{right})$, represents the acceleration signals.
The human motion output $\mathbf{M}=\{\mathbf{m}_t\}_{t=0}^T$ is a sequence of pose $\mathbf{m}_t$, and $\mathbf{m}_t =(\mathbf{p}^0_t, \mathbf{q}^0_t, \mathbf{q}^1_t,..., \mathbf{q}^J_t)  \in \mathbb{R}^{3J+3}$ is a concatenated vector of root translation $\mathbf{p}^0 \in \mathbb{R}^{3}$ defined in a world coordinate and local joint rotations of $J$ joints where the rotation of $j$-th joint is represented as $\mathbf{q}^j \in so(3)$. 

We first apply an off-the-shelf monocular SLAM~\cite{teed2021droid} with the egocentric video $\mathbf{I}$. Given the original world coordinate arbitrarily defined from the SLAM model, we redefine the world coordinate via an alignment procedure described in Sec.~\ref{sec:preprocess}, such that the scale of the world coordinate reflects the metric scale (in meter) and the negative z-axis is aligned to the gravity direction. From the SLAM, we obtain the 3D pointcloud $\mathbf{W}$ of the environment and camera pose at each time  $\mathbf{C} = \{\mathbf{C}_t\}_{t=0}^{T}$, defined in the world coordinate. Then, given the camera poses $\mathbf{C}$, we compute the head pose $\mathbf{H} = \{\mathbf{H}_t\}_{t=0}^T$, where $\mathbf{H}_t \in SE(3)$ is the location and orientation of the head joint. The head poses are directly leveraged into the estimation pipeline with sensor inputs, playing a key role in disambiguating motions during estimation. Notably, in non-flat spaces as in Fig.~\ref{fig:input_vis_origin}, the z-directional component of $\mathbf{H}_t$ defined in the world coordinate does not necessarily reflect the metric height of the person because the height should be defined from the actual floor the person is currently standing. Thus, we compute the height of the head $\mathbf{H}_t$ by tracking and updating the floor level $f_t \in \mathbb{R}$ at time $t$ accordingly. 

We build a transformer-based regression model coupled with a floor-level update module to estimate the full-body human motion from the cues of estimated head pose and wrist IMU signals: $\mathbf{M}_{reg} = \mathcal{F}_{est} (\mathbf{H}, \mathbf{W}, \mathbf{S}_{IMU})$.
While our initial full-body estimation $\mathbf{M}_{reg}$ is already compelling, there exist fundamental ambiguities due to the extreme sparsity of our input signals. 
To address this issue, we leverage the visual cues captured by the head-mounted view $\mathbf{I}$, where the hand position and the interactions with the environments are observed. While available only occasionally, we demonstrate the use of such visual cues can enhance the motion capture quality.
Going one step further, we also consider multi-people capture scenarios where each individual wears our lightweight system. Assuming the visual signals among people can be shared, we can additionally leverage the occasional ``third-person'' views from other people for the target person's motion capture. 
To fuse all of these available signals for more accurate motion estimation, we build a motion optimizer module $\mathcal{F}_{opt}$ where $\mathbf{M} = \mathcal{F}_{opt} (\mathbf{M}_{reg}, \mathbf{\Phi})$. Optimizing the human motion is done in a spatiotemporal manifold space~\cite{holden2016deepsynthesis}, where $\mathbf{M}_{reg}$ is the initial output from our motion regressor module and $\mathbf{\Phi}$ is the visual cues extracted from head-mounted camera or third-person views from other users' cameras. For $\mathbf{\Phi}$, we mainly use the 2D keypoint cues estimated from images in egocentric scenarios and 3D pose estimation results in multi-person scenarios.

\subsection{Pre-processing Sensor Inputs}
\label{sec:preprocess}
\noindent \textbf{Head Trajectory from Monocular SLAM.} From egocentric video $\mathbf{I}$, we first apply DROID-SLAM~\cite{teed2021droid}, noted for its accuracy and robustness compared to classical SLAM systems,
to estimate camera trajectory $\mathbf{C}$ and reconstruct the 3D pointcloud $\mathbf{W}$. 
In some cases, however, outliers may exist in camera pose estimation 
due to insufficient textures in the scenes or blurs, which may negatively affect our body pose estimation quality. As a way to filter out outliers, we compute the temporal acceleration between camera movements and detect erroneous camera pose estimation when the acceleration values are over a certain threshold. We fill in the missing camera poses via linear interpolation. 

\noindent \textbf{Aligning the Coordinate for Camera and IMU Sensors.} 
For preprocessing, we align the arbitrarily defined original coordinate from SLAM into our desired real-world coordinate with metric scale and the negative $z$-axis to be aligned to the gravity direction. Additionally, we also calibrate the orientation of IMUs to have the common orientation axes aligned to the gravity direction. This is done via two steps: (1) aligning IMU sensors to real-world coordinates, and (2) aligning the coordinate from SLAM to IMU coordinates. 

Because the camera center $\mathbf{C}$ may not be necessarily the same as the head joint location, we compute the fixed transformation $T_{head}^{cam}$ to transform the camera pose into the head pose, resulting in $\mathbf{H}$ defined w.r.t world coordinate.
$T_{head}^{cam}$ is computed by approximating the camera location in a surface point of SMPL mesh.
Refer to the supp. mat. for the details of the alignment protocol and time synchronization.

\subsection{Motion Estimation}
\label{sec:motion_est}

\noindent  \textbf{Pre-Processing Input Signals.}
The motion estimator module $\mathcal{F}_{est}$ is a transformer-based network to estimate motion and foot contacts from head trajectory $\mathbf{H} = \{\mathbf{H}_t\}_{t=0}^{T}$ and IMU sensor signals $ \mathbf{S}_{IMU} = \{ \mathbf{R}, \mathbf{A} \}$.
We input the data into the network in a sliding temporal window manner with length $N$. For every window, we normalize the input to be local to the head coordinate $\mathbf{H}_{\tau}$ at $\tau=0$ which is the first frame of the current window by applying a transformation matrix $T_{\mathbf{H}_\tau}^{w}$ that transforms the world coordinate to the coordinate of $\mathbf{H}_{\tau=0}$. 
The normalized cues are denoted with the hat symbol: $\mathbf{\hat{H}} = \{\mathbf{\hat{H}}_\tau\}_{\tau=0}^{N}$ , $\mathbf{\hat{R}} = \{ \mathbf{\hat{R}}_\tau \}_{\tau=0}^{N}$, $\mathbf{\hat{A}} = \{ \mathbf{\hat{A}}_\tau \}_{\tau=0}^{N}$, and $\mathbf{\hat{m}}_\tau = \{ \mathbf{\hat{p}}^0_\tau, \mathbf{\hat{q}}^0_\tau, \mathbf{q}^1_\tau, ... \mathbf{q}^J_
\tau\}$\footnote{In the normalized output pose, the local joint rotations $\mathbf{q}_t^1, ... , \mathbf{q}_t^J$ is identical to the original pose $\mathbf{m}_t$ as local joint rotations are coordinate invariant, while the root orientation $\mathbf{q}_t^0$ is changed.}.
The normalized output motion is recovered by applying $T^{\mathbf{H}_\tau}_{w}$, or $(T_{\mathbf{H}_\tau}^{w})^{-1}$.

\begin{figure}
\centering
\includegraphics[width=0.9\linewidth, trim={0 0cm 0 0}]{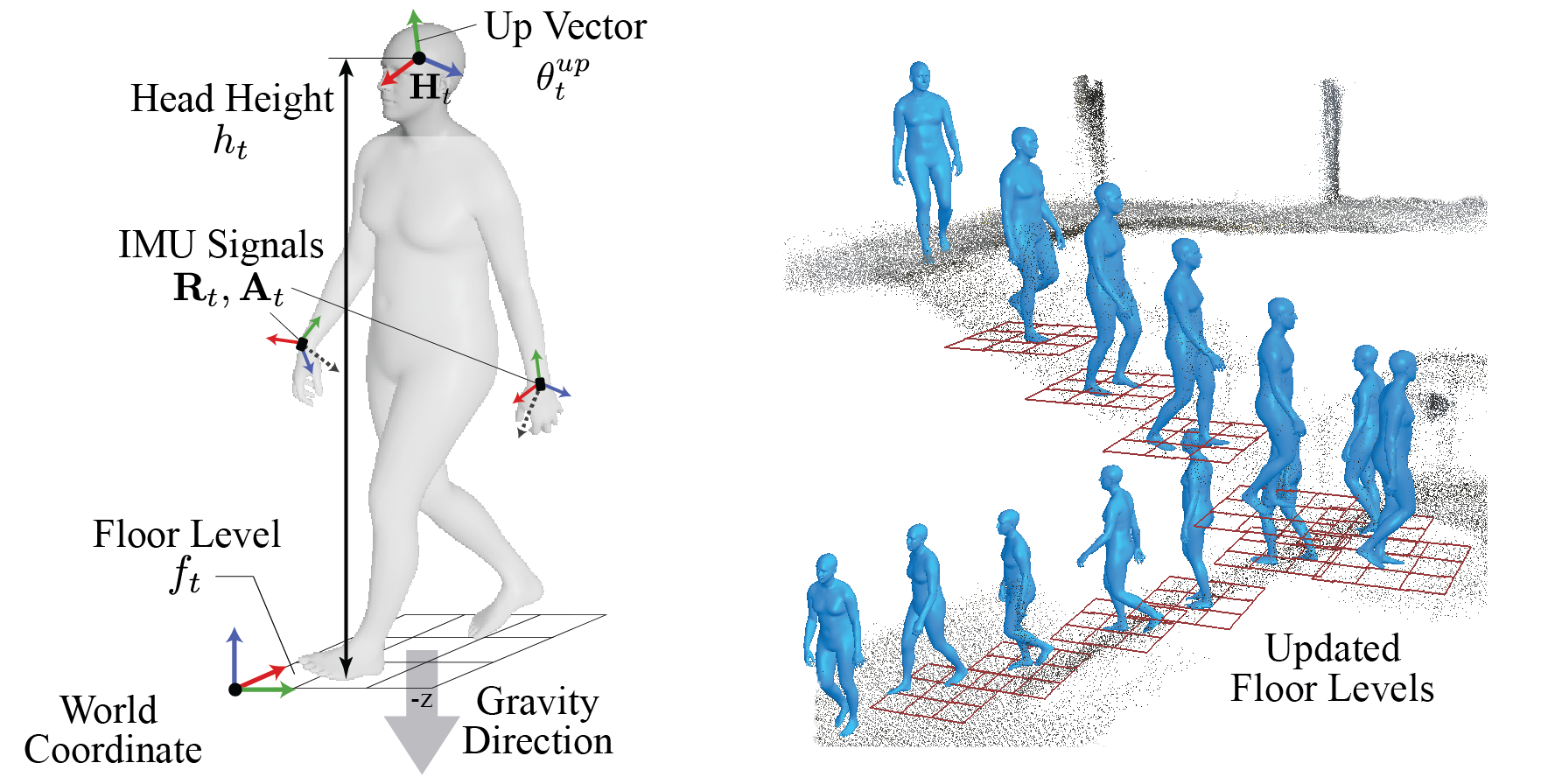}
    \caption{(a) Visualization of input signals. (b) Visualization of the updated floor levels $f_t$.}
\label{fig:input_vis_origin}
\vspace{-0.8cm}
\end{figure}

Normalizing every cue to the first frame head coordinates may miss out on important absolute cues critical to defining human motions. For example, a person standing still and sitting still would have identical input when normalized.
Thus, we furthermore include absolute head height (z-axis value) $h_t \in \mathbb{R}$, and the head up-vector $\theta^{up}_t \in \mathbb{R}^{3}$ defined in world coordinate. 
To accurately define head height $h$, we introduce an algorithm to update floor level $f_t$ at time $t$ based on network output (motion and foot contact) obtained until time $t-1$ with 3D pointcloud $\mathbf{W}$. Once the $f_t$ is estimated, as described in Fig.~\ref{fig:input_vis_origin}, the head height $h_t$ is computed based on the floor level, or $h_t = \mathbf{H}_z - f_t$.  
The input components are concatenated into 
$\{\mathbf{x}_\tau\}_{\tau=0}^N = \{\mathbf{\hat{H}}, \mathbf{\hat{R}}, \mathbf{\hat{A}}, h_\tau, \theta^{up}_\tau\} \in \mathbb{R}^{N \times 31} $. Before concatenating, rotations are converted into 6D representations~\cite{zhou20196d}. 

\noindent \textbf{Network Architecture.}
We formulate the estimation as a sequence-to-sequence (seq2seq) problem to effectively incorporate temporal information to address the ambiguities arising from the sparsity of sensor inputs.
Following the previous work~\cite{jiang2022avatarposer}, we utilize Transformer encoder models and their self-attention mechanism to efficiently capture intricate relationships in the time-series data.

Different from the previous work~\cite{jiang2022avatarposer} that directly maps input sequence $\{\mathbf{x}_\tau\}_{\tau=0}^N$ to the output motion $\mathbf{\hat{m}}_\tau$, we adopt a multi-stage method as in~\cite{yi2021transpose} by introducing end-effector positions as intermediate representations. 
Specifically, we separate the system $\mathcal{F}_{est}$ into two submodules $\mathcal{F}^{end}$ and $\mathcal{F}^{body}$ (we drop the subscript `$est$' on these submodule names for brevity). The first submodule $\mathcal{F}^{end}$ takes $\{\mathbf{x}_\tau\}_{\tau=0}^N$ as input and generates end-effector positions (hands and feet)
$\{ \mathbf{x}_\tau^{mid} \}_{\tau=0}^{N} \in \mathbb{R}^{N \times 12}$. 
The input $\{\mathbf{x}_\tau\}$ and $\{ \mathbf{x}_\tau^{mid} \}$ are fed into submodule $\mathcal{F}^{body}$.
The $\mathcal{F}^{body}$ regresses the output motion $\{ \mathbf{\hat{m}}_\tau \}_{\tau=0}^{N}$ and also foot contact $\{\mathbf{c}_\tau\}_{\tau=0}^N$ where $\mathbf{c}_\tau = \{ c^{lf}_\tau, c^{rf}_\tau \}$. $c^{lf}_\tau$ and $c^{rf}_\tau$ indicate the contact probability of the left and right foot at time $\tau$.  

\begin{figure}
\centering
\includegraphics[width=0.95\linewidth, trim={0 0.2cm 0 0}]{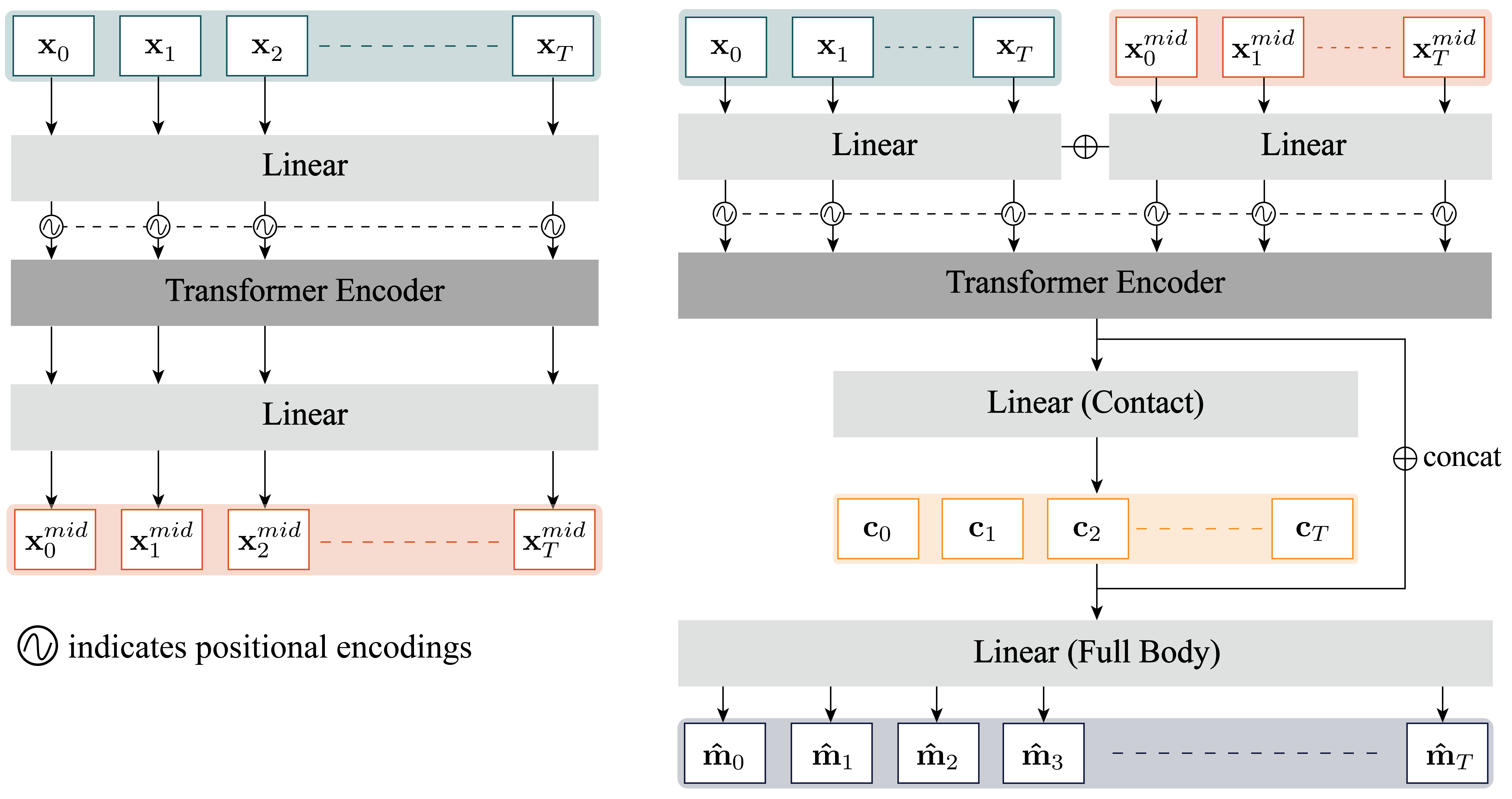}
    \caption{Network architecture of submodules $\mathcal{F}^{end}$ (left) and $\mathcal{F}^{body}$ (right) of $\mathcal{F}_{est}$.}
    \label{fig:network_arch}
\vspace{-0.5cm}
\end{figure}

Both submodules are based on Transformer encoder models with linear embedding layers to project the input vector into continuous embeddings in the first part of the networks. For $\mathcal{F}^{body}$ where $\{\mathbf{x}_\tau^{i}\}$ and $\{\mathbf{x}_\tau^{mid}\}$ are concatenated as input, there are two separate linear layers and the linear embeddings are concatenated and fed into the Transformer encoder. In submodule $\mathcal{F}^{end}$, the features generated by the Transformer encoder are converted into mid-representations $\{\mathbf{x}_\tau^{mid}\}$ by a 2-layer MLP. In submodule $\mathcal{F}^{body}$, the output of the Transformer encoder is first converted into foot contact probabilities $\{\mathbf{c}_\tau\}$. The contact probability and Transformer encoder output are concatenated and converted to output motion $\{\mathbf{\hat{m}}_\tau\}$. 
The network architecture for both submodules are in Fig.~\ref{fig:network_arch}.

\noindent \textbf{Updating Floor Levels.}
Among the input $\{ \mathbf{x}^{i}_\tau\}$, the height cue $h$ represents the metric height of the person's head joint, which should be computed from the actual floor the character stands on.
Assuming large-scale indoor and outdoor environments (e.g., multiple floors with stairs, and uphills), the actual floor the character stands on is not necessarily the same as the $z=0$ plane in the world coordinate defined in the alignment procedure (Sec.~\ref{sec:preprocess}).
Therefore, the floor level should be updated based on the current character status and the environment. 

The key idea of updating floor levels is to track the foot contact $\{\mathbf{c}_t\}_{t=0}^{t-1}$. As foot contact is a form of interaction between the human and the floor, 
the foot position during contact can be considered as the floor level. 
From the contact estimation $\{\mathbf{c}_t\}_{t=0}^{t-1}$ obtained by $\mathcal{F}^{body}$, we find the latest time frame $t_{m}$ where the foot contact occurs in either foot with a confidence value above a certain fixed threshold, $c^f_{t_{m}} > \lambda$. Given the corresponding foot joint location $\mathbf{p}_{t_{m}}^{f}$ in contact, the floor point can be obtained by projecting $\mathbf{p}_{t_{m}}^{f}$ to the 3D scene pointcloud $\mathbf{W}$, where we simply find nearby points and take the mean of their $z$ values as $f_t$.
Visual examples of updated floors are shown in Fig.~\ref{fig:input_vis_origin} (b)\footnote{While we only keep the floor height $f_t$, the 3D floor planes in Fig.~\ref{fig:input_vis_origin} are shown for visualization purposes using root's xy positions at time $t$.}.

\noindent \textbf{Training.}
The submodules $\mathcal{F}^{end}$ and $\mathcal{F}^{body}$ are trained end-to-end, with the following loss terms
$\mathcal{L}_{pos}$, $\mathcal{L}_{rot}$, $\mathcal{L}_{root}$, $\mathcal{L}_{mid}$, $\mathcal{L}_{contact}$, $\mathcal{L}_{footvel}$, and $\mathcal{L}_{cons}$.
The loss term $\mathcal{L}_{root}$ and $\mathcal{L}_{rot}$ are:
\begin{equation}
    \begin{split}
        \mathcal{L}_{root} &= \|\mathbf{\hat{p}}^0 - \mathbf{\hat{p}}_{GT}^{0}\|  + \| \mathbf{\hat{q}}^0 - \mathbf{\hat{q}}^0_{GT}\|  \\ 
        \mathcal{L}_{rot} &= \sum_{j \in J}\| \mathbf{q}^j - \mathbf{q}^j_{GT}\|.
    \end{split}
\end{equation}
The terms $\mathcal{L}_{root}$ and $\mathcal{L}_{rot}$ enforces the output motion $\mathbf{\hat{m}}_t = \{ \mathbf{\hat{p}}_t^{0}, \mathbf{\hat{q}}_t^{0}, ... , \mathbf{q}_t^{J} \}$ to follow the given ground truth data.

The loss term $\mathcal{L}_{pos}$ penalizes the error accumulating along the kinematic chain by considering position values of each joint, and is measured by the weighted difference of joint positions.
\begin{equation}
    \mathcal{L}_{pos} = \sum_{j \in J} w_j \| \mathbf{\hat{p}}^j - \mathbf{\hat{p}}^j_{GT} \| 
\end{equation}
where $\mathbf{\hat{p}}^j$ indicates position of joint $j$, which is obtained by forward kinematics operation.
 In case of foot joints, 
 $\mathcal{L}_{footvel} = \| \mathbf{\hat{v}}^{foot} - \mathbf{\hat{v}}^{foot}_{GT}  \|$ is additionally computed to penalize foot slip artifacts and unnatural foot movements. $\mathbf{\hat{v}}^{j}$ indicates velocity of joint $j$.

$\mathcal{L}_{mid}$ and $\mathcal{L}_{cons}$ are terms to jointly train the submodule $\mathcal{F}^{end}$ and $\mathcal{F}^{body}$. 
\begin{equation}
    \begin{split}
        \mathcal{L}_{mid} &= \sum_{j \in end} \| \mathbf{\hat{p}}^j_{mid} - \mathbf{\hat{p}}^j_{GT} \| \\
        \mathcal{L}_{cons} &= \sum_{j \in end} \| \mathbf{\hat{p}}^j - \mathbf{\hat{p}}^j_{mid} \|
    \end{split}
\end{equation}
where $\mathbf{\hat{p}}^j_{mid}$ indicates the output of $\mathcal{F}^{end}$ (end-effector positions).
$\mathcal{L}_{mid}$ directly trains the $\mathcal{F}^{end}$ module to generate accurate intermediate representations $\mathbf{x}_\tau^{mid}$.
The consistency loss $\mathcal{L}_{cons}$ enforces the consistency between mid-representations and the output motion.
Intuitively, the end-effector joint positions derived from the output motion $\mathbf{\hat{m}}_t$ should be consistent with the intermediate output $\mathbf{x}_\tau^{mid}$ which is also the positions of end-effector joints.
The two loss terms enforce $\mathcal{F}^{end}$ to provide reliable additional information 
to submodule $\mathcal{F}^{body}$, and enforce $\mathcal{F}^{body}$ to reflect the provided mid-representation to the final output.

$\mathcal{L}_{contact} = BCELoss(\mathbf{c}, \mathbf{c}_{GT})$ is to train the model to estimate contact probabilities $\mathbf{c}$.
Here the subscript GT indicates ground truth while time $\tau$ is omitted for convenience.

\subsection{Motion Optimization With Visual Cues}
\label{sec:motion_opt}
\noindent \textbf{Generating Visual Cues From Images.} 
Since we only use upper body sensors, fundamental ambiguities exist on the lower body joints. For example, sitting and crossing one's legs cannot be easily distinguishable from other sitting postures. Also, different from VR-based systems, the absence of explicit positional data in raw IMU sensors can lead to uncertainties in determining hand positions.
Although such ambiguities can be resolved in dynamic body movements (e.g., locomotion) by taking the dynamics into account via temporal information, our daily activities (e.g., manipulation) may often contain subtle and sophisticated motions with minimum movements. 
As a solution to enhance the motion capture accuracy, we present a method to leverage the visual cues from the head-mounted camera, based on the observation that it is common to look at one's hands and the object during typical hand-involved interactions. 
The idea can be extended into multi-people scenarios, where we can assume multiple people wear our motion capture systems and their visual cues are shared with each other. As demonstrated in our experiments, a person's egocentric view can be treated as occasionally available sparse 3rd person views which can be used as informative additional visual cues.
Motivated by this idea, we leverage the RGB information of the egocentric video $\mathbf{I}$.
The visual cues $\mathbf{\Phi} = [\phi_{E}, \phi_{T}]$ represent additional information of the human pose. In single-person egocentric setting,  $\phi_{E} = \{\bm{\mu}_t^j \}$ where $\bm{\mu}_t^j \in \mathbb{R}^2$, indicates the 2D location of joint $j$ in 2D image coordinates. Joint $j$ is either left or right wrist, or both in practice.
In the case of multi-person capture scenarios, we can leverage a monocular 3D human pose estimation approach from 3rd person views of other individuals, which can be used during our motion optimization. The estimated 3D pose is denoted as $\phi_{T} = \{ \bm{\tau}_t^{\overrightarrow{BA}} \}$, where $\bm{\tau}_t^{\overrightarrow{BA}} = (\mathbf{q}_t^1 ... \mathbf{q}_t^j)^A$ indicates local joint rotations (excluding the root) of person $A$ estimated via $\mathbf{I}_t^B$, or image taken by person $B$ at time $t$. We use off-the-shelf models~\cite{lugaresi2019mediapipe, goel20234dhumans} to obtain visual cues $\phi_E$ and $\phi_T$. 

\noindent \textbf{Manifold-Based Motion Optimization.}
The optimized motion $\mathbf{M}$ should not only fulfill the occasionally detected cues $\mathbf{\Phi}$, but it should maintain the semantics and naturalness of $\mathbf{M}_{reg}$, preserving spatiotemporal coherency. 
Instead of directly optimizing the motion output from $\mathbf{M}_{reg}$ to $\mathbf{\Phi}$, we utilize the motion manifold-based method demonstrated in~\cite{holden2016deepsynthesis, lee2023lama}. As the motion manifold is learned to preserve spatio-temporal correlation of human movements, optimizing within this manifold space can enforce the optimized motion $\mathbf{M}$ to maintain its naturalness while fulfilling the target cues $\mathbf{\Phi}$. 
Especially in the case of egocentric visual cue $\phi_E$ where only 2D joint cues are available, optimizing within this manifold space helps to maintain naturalness in the optimized motion despite the ambiguity of the cue.

To build motion manifolds we use convolutional autoencoders~\cite{holden2016deepsynthesis}, compressing motion sequences into corresponding latent vectors.
For training, the motion sequence $\mathbf{X} \in \mathbb{R}^{T\times137}$ is represented as a concatenated vector of foot contact, root translation, and joint rotations. The root translation and orientation are normalized based on the first frame of the motion sequence. The rotation values are converted into 6D representations~\cite{zhou20196d} for network learning.
The encoder $E$ and decoder module $E^{-1}$ are trained based on reconstruction loss, or $ \mathcal{L}_{recon} = ||\mathbf{X} - E^{-1}( E\left(\mathbf{X}) \right) ||^2$.

Motion optimization is done by searching an optimal latent vector among the manifold. 
From the initial latent vector $\mathbf{z} = E(\mathbf{X})$, where $\mathbf{X}$ is computed from $\mathbf{M}_{reg}$,  the optimal latent vector $\mathbf{z}^*$ is found by minimizing loss $\mathcal{L}= \mathcal{L}_{vis} + \mathcal{L}_{reg} + \mathcal{L}_{contact}$.\footnote{Here the superscript $\overrightarrow{BA}$ and $A$ indicating humans in $\bm{\tau}$ is omitted for convenience, and is replaced with joint indicators. $ \bm{\tau}_t^{j} = \bm{\tau}_t^{\overrightarrow{BA}}$ of joint $j$. } $\mathcal{L}_{vis}$ enforces to meet the provided visual cues:
\begin{equation}
    \begin{split}
        \mathcal{L}_{vis} &= \begin{cases}
        || \bm{\mu}_t^j - \pi(\mathbf{C}_t, \mathbf{p}_t^j) || &\bm{\mu}_t^j \in \phi_E \\
        \sum_{j} w_j || \bm{\tau}_t^{j} - \mathbf{q}_t^j || & \bm{\tau}_t^{j} \in \phi_T.
\end{cases} \\
    \end{split}
\end{equation}
The function $\pi(\mathbf{C}_t, \mathbf{p}_t^j)$ indicates the projection of 3D joint position $\mathbf{p}_t^j$ with camera $\mathbf{C}_t$. 
$\mathcal{L}_{reg}$ is for regularization and $\mathcal{L}_{contact} = \mathbf{c}_t \cdot \|\dot{\mathbf{p}}_t^{foot}\|$ penalizes foot slip when the foot is in contact.
Finally, the optimized motion $\mathbf{M}$ is derived from the optimized latent vector $\mathbf{z}^*$ by $E^{-1}(\mathbf{z}^*)$.

\section{Experiments}
\label{sec:experiment}

\subsection{Experimental Setup}

\noindent\textbf{Synthesizing IMU Data from Mocap.}
To build training data from motion datasets, we synthesize IMU signals using wrist poses, following the protocol in~\cite{yi2021transpose, yi2022pip, jiang2022tip}.

\noindent\textbf{Processing IMU Signals from Smartwatches.}
For demonstrations with real-world IMU data using smartwatches, we apply filtering to reduce noise as in~\cite{jiang2022tip} beforehand.

\noindent\textbf{Evaluation Metrics.} 
We adopt the following metrics for quantitative experiments.

$\sbullet$ \textbf{Mean Per Joint Position Error (MPJPE):} MPJPE represents the average position error per joint. 
($cm$)

$\sbullet$ \textbf{Mean Per Joint Velocity Error (MPJVE):} MPJVE measures the average velocity error across joints. ($cm/s$)

 $\sbullet$ \textbf{Jitter:} Jitter indicates how smooth the motion is, and is measured by acceleration changes over time averaged by body joints. Here we present the relative ratio between the jitter of the predicted motion and the ground truth.

\noindent\textbf{Baselines.} 
Since we are the first to estimate full body motion from a head-mounted camera and two IMU sensors, no direct baselines exist. Thus, we compare our estimation module $\mathcal{F}_{est}$ with previous SOTAs based on (1) 6 IMUs (full body)~\cite{yi2022pip, jiang2022tip}, and (2) VR devices (upper body, 6 DoF)~\cite{du2023agrol}, which pursue similar goals with more information than our setup through additional sensors or 6DoF hand measurements. 
In the context of wearable and outdoor motion capture, we consider IMU-based methods as our main competitors.
However, for completeness, we additionally compare with the VR device-based methods due to their technical similarity to our approach of estimating motion from upper body sensors. 
Note that in this comparison we assume fixed flat floor planes since previous approaches cannot handle varying floor levels.

$\sbullet$ \textbf{IMU Baselines (Full Body):} 
We compare ours with \textbf{PIP}~\cite{yi2022pip} and \textbf{TIP}~\cite{jiang2022tip}.
Both estimate motion from 6 IMU sensors on the full body; head, wrists, pelvis, and legs. In the supp. mat. we also compare with \textbf{EgoLocate}~\cite{yi2023egolocate} which leverages an additional head-mounted camera on top of the full-body IMU sensors for global translation correction.

$\sbullet$ \textbf{VR Baselines (Upper Body):} 
We consider \textbf{AGRoL}~\cite{du2023agrol}, the state-of-the-art method for tracking full body motion from 3 6DoF (position and orientation) sensor signals from the head and both wrists, as a baseline. We compare with two versions of AGRoL, the original version where the 6DoF values are given for all 3 sensors and a modified version where the position values of the wrist are replaced with acceleration as in IMU sensors.

$\sbullet$ \textbf{Ablative Baselines:} We perform ablation studies on the floor level update in $\mathcal{F}_{est}$ to present the contribution of the algorithm on scenarios with large areas with drastic floor changes such as walking down the stairs from the third to the second floor of a building. We furthermore compare the output of $\mathcal{F}_{est}$ only and with $\mathcal{F}_{opt}$ to demonstrate the contribution of egocentric video to enhance motion estimation.

\subsection{Comparison with IMU-Based Baselines}
\noindent\textbf{Dataset.} 
We follow the previous approaches~\cite{yi2022pip, jiang2022tip} by training the models on AMASS~\cite{mahmood2019amass} dataset and evaluating on the TotalCapture~\cite{trumble2017totalcapture} dataset with real IMU data. 
\footnote{While previous approaches use DIP dataset~\cite{huang2018dip} as well, we exclude this in our test, because DIP does not include global translations and we cannot generate head positions from the dataset. DIP is much smaller than AMASS and consists of less than 5\% of the whole training dataset.}

\noindent\textbf{Results.}
As shown in Table~\ref{tab:baseline_imu_quant},
despite only utilizing sensors on the upper body, our method shows comparable or better performances compared to 6 IMU-based setups. 
In the root-relative position errors (r.MPJPE) which compare pose estimation quality ignoring root drifts, our approach outperforms the baselines. Notably, although PIP~\cite{yi2022pip} explicitly correct pose with physics-based optimization, the performance of ours is comparable without such explicit corrections.
In other metrics where root drift errors are also considered (MPJPE, Root PE), our method shows much better performance with significant decreases in root drift compared to baselines 
(MPJPE decreases by 87.7\% for PIP and 89.8\% for TIP; Root PE by 90.5\% and 92.1\%, respectively). 
For comparison with EgoLocate, results in supp. Table 2 shows our superior performance, especially in root-related position error terms (MPJPE, Root PE), despite the reduced number of sensors. 
These results demonstrate that incorporating head pose directly in estimation can improve performance and mitigate root drift issues without any additional localization or correction as done in previous approaches.

\begin{table}
    \footnotesize
    \centering
    \begin{tabular}{c|c|c|c|c|c}
    \toprule
        Method & r.MPJPE & MPJPE & MPJVE & Root PE & Jitter  \\
    \midrule
    PIP~\cite{yi2022pip} & \underline{4.40} & \underline{34.69} & \textbf{19.51} & \underline{34.15} &  \textbf{0.95} \\
        TIP~\cite{jiang2022tip} & 4.88 & 41.79 & 36.26  & 41.18 & 16.52 \\
        Ours ($\mathcal{F}_{est}$)  & \textbf{3.77} &  \textbf{4.27} & \underline{19.56} &\textbf{3.25} & \underline{5.23} \\
    \bottomrule
    \end{tabular}
    \caption{Comparison with full body IMU-based methods on the TotalCapture dataset.}
    \label{tab:baseline_imu_quant}
        \vspace{-0.25cm}
\end{table}

\begin{table}
    \footnotesize
    \centering
    \begin{tabular}{c|c|c|c|c|c}
    \toprule
         Dataset & Method & MPJPE (r.) & MPJVE & Root PE & Jitter \\
    \midrule  
  AMASS & AGRoL~\cite{du2023agrol} & \textbf{4.58 (4.89)} & \underline{19.31} & \textbf{4.15} & \underline{2.60}  \\
          & AGRoL* & 6.53 (6.18) & 24.99 & 5.33 & 3.49 \\
          & Ours ($\mathcal{F}_{est}$) & \underline{5.20 (4.95)} & \textbf{17.00} & \underline{4.36} & \textbf{1.64} \\
    \midrule  
        HPS & AGRoL~\cite{du2023agrol} & \underline{31.47 (27.64)} & \underline{258.92} & \underline{25.76} & \underline{40.83} \\
        & AGRoL* & 33.46 (27.69) & 389.83 & 28.22 & 62.06 \\
          & Ours ($\mathcal{F}_{est}$) & \textbf{8.65 (8.24)} & \textbf{21.42} &  \textbf{6.97} & \textbf{1.84}  \\
    \bottomrule
    \end{tabular}
    \caption{Comparison with upper body VR-based methods on AMASS and HPS datasets. AGRoL with an asterisk is the modified version where wrist position are replaced with accelerations. (r.) indicates root-relative position errors.}
    \label{tab:baseline_comparison_vr}
    \vspace{-0.6cm}
\end{table}

\subsection{Comparison with VR-Based Baselines}
\noindent\textbf{Dataset.} For quantitative comparison, we 
generate train/test/validation split from a subset of AMASS dataset (CMU~\cite{cmumocap}, HDM05~\cite{muller2007hdm05}, BMLMovi~\cite{ghorbani2020movi}). We first evaluate our motion estimation module and AGRoL on the testing split of the AMASS dataset. Additionally, we also demonstrate the generalization ability of the modules by testing on the HPS~\cite{guzov2021hps} dataset. Regarding HPS dataset, we use the motion optimization result~\cite{guzov2021hps} as ground truth. We randomly select 10 sequences in 6 scenes for comparison.

\noindent\textbf{Results.} 
As demonstrated in Table~\ref{tab:baseline_comparison_vr},
our method shows on-par performance with the original AGRoL, even though in AGRoL the positions of both hands are provided.
Compared with the modified AGRoL where wrist positions are replaced with accelerations, our module $\mathcal{F}_{est}$ outperforms it in all of the metrics. 
Furthermore, our module $\mathcal{F}_{est}$ shows a notable performance gap compared to AGRoL on the HPS dataset which includes motions in large-scale scenes. This is expected as AGRoL focuses on full body motion estimation in a VR-optimized setting, not in large-scale scenes.

\subsection{Ablation Studies}
\noindent\textbf{Setup.}
For quantitative evaluation, we capture with our setup and with XSens MVN Link~\cite{xsens} together, where we use XSens as ground truth.
More details are in supp. mat.

\begin{figure}
    \centering    \includegraphics[width=0.95\linewidth, trim={0cm 0cm 0cm 0cm}]{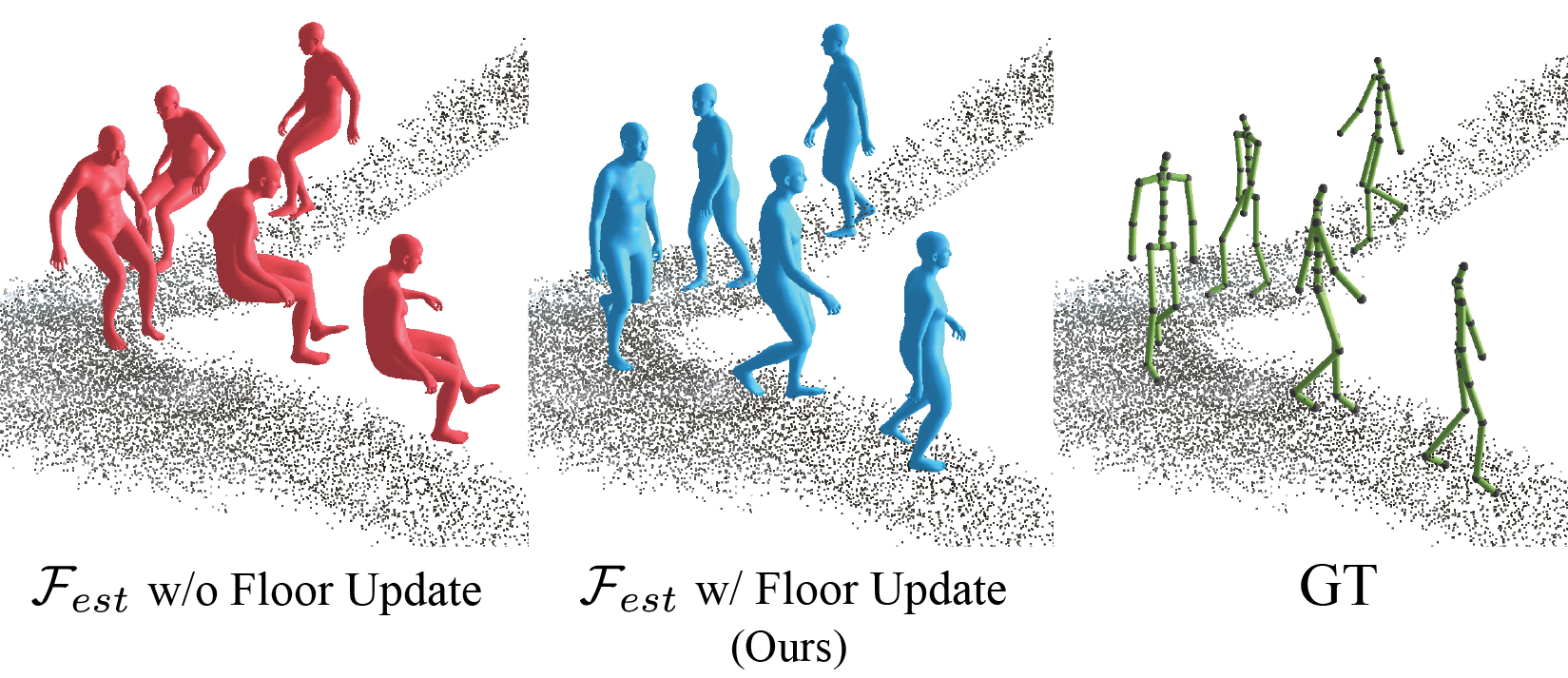}
    \caption{Comparison with $\mathcal{F}_{est}$ without floor update (left) and with floor update (right). The floor update algorithm corrects head height for estimating accurate poses.}
    \label{fig:floor_abl}
        \vspace{-0.2cm}
\end{figure}

\noindent\textbf{Ablation on Motion Estimation.} We demonstrate the contribution of the floor-level updating algorithm by capturing a sequence of walking downstairs.
As seen in Table~\ref{tab:floor_ablation} and Fig.~\ref{fig:floor_abl}, without floor update (left) the head height is misleading and consequently, the module generates inaccurate poses. However, by accordingly updating the floor levels (right) the head height is adjusted so the network accurately estimates the motion of walking down the stairs.

\begin{table}
    \footnotesize
    \centering
    \begin{tabular}{c|c|c|c}
    \toprule
        Method & Upper PE & Lower PE & Full PE \\
    \midrule
        $\mathcal{F}_{est}$ w/o Floor Update & 9.19 & 16.58 & 12.35 \\
       $\mathcal{F}_{est}$ w/ Floor Update (Ours) & \textbf{6.04} & \textbf{7.19} & \textbf{6.53} \\ 
    \midrule
    \end{tabular}
    \caption{Quantitative comparison with and $\mathcal{F}_{est}$ without floor level update.}
        \vspace{-0.4cm}
    \label{tab:floor_ablation}
\end{table}

\begin{table}
    \footnotesize
    \begin{tabular}{c|c|c|c}
    \toprule
        Method & Right Hand PE & Right Arm PE & Upper PE \\
    \midrule
        $\mathcal{F}_{est}$ & 11.91 & 9.72 & 8.04 \\
       $\mathcal{F}_{est} + \mathcal{F}_{opt}$ (Ours) & \textbf{5.55} & \textbf{5.97} & \textbf{6.65} \\ 
    \midrule
    \end{tabular}
    \caption{Quantitative comparison with $\mathcal{F}_{est}$ and $\mathcal{F}_{est} + \mathcal{F}_{opt}$ optimized with egocentric-view visual cues.}
    \label{tab:edit_ablation}
    \vspace{-0.4cm}
\end{table}

\begin{figure}
\centering    \includegraphics[width=0.93\linewidth, trim={0cm 0.2cm 0cm 0.3cm}]{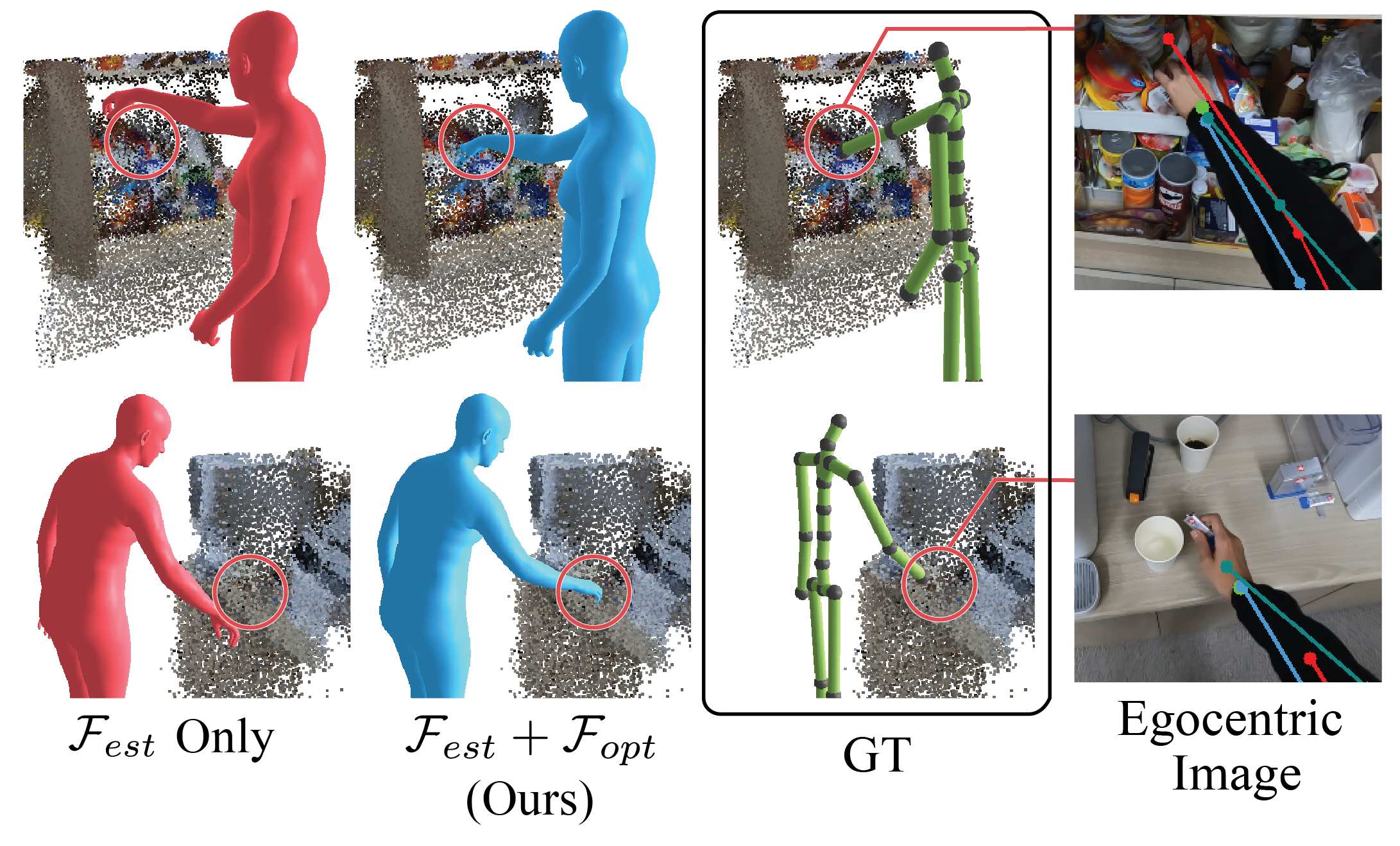}
    \caption{Comparison with $\mathcal{F}_{est}$ (left) and $\mathcal{F}_{est} + \mathcal{F}_{opt}$ (right). The optimization module $\mathcal{F}_{opt}$ corrects the arm and hand pose based on the 2D hand keypoint detected in the egocentric image.}
    \label{fig:edit_abl}
    \vspace{-0.2cm}
\end{figure}

\begin{figure}
\centering
\includegraphics[width=0.95\linewidth, trim={0 0.1cm 0 0}]{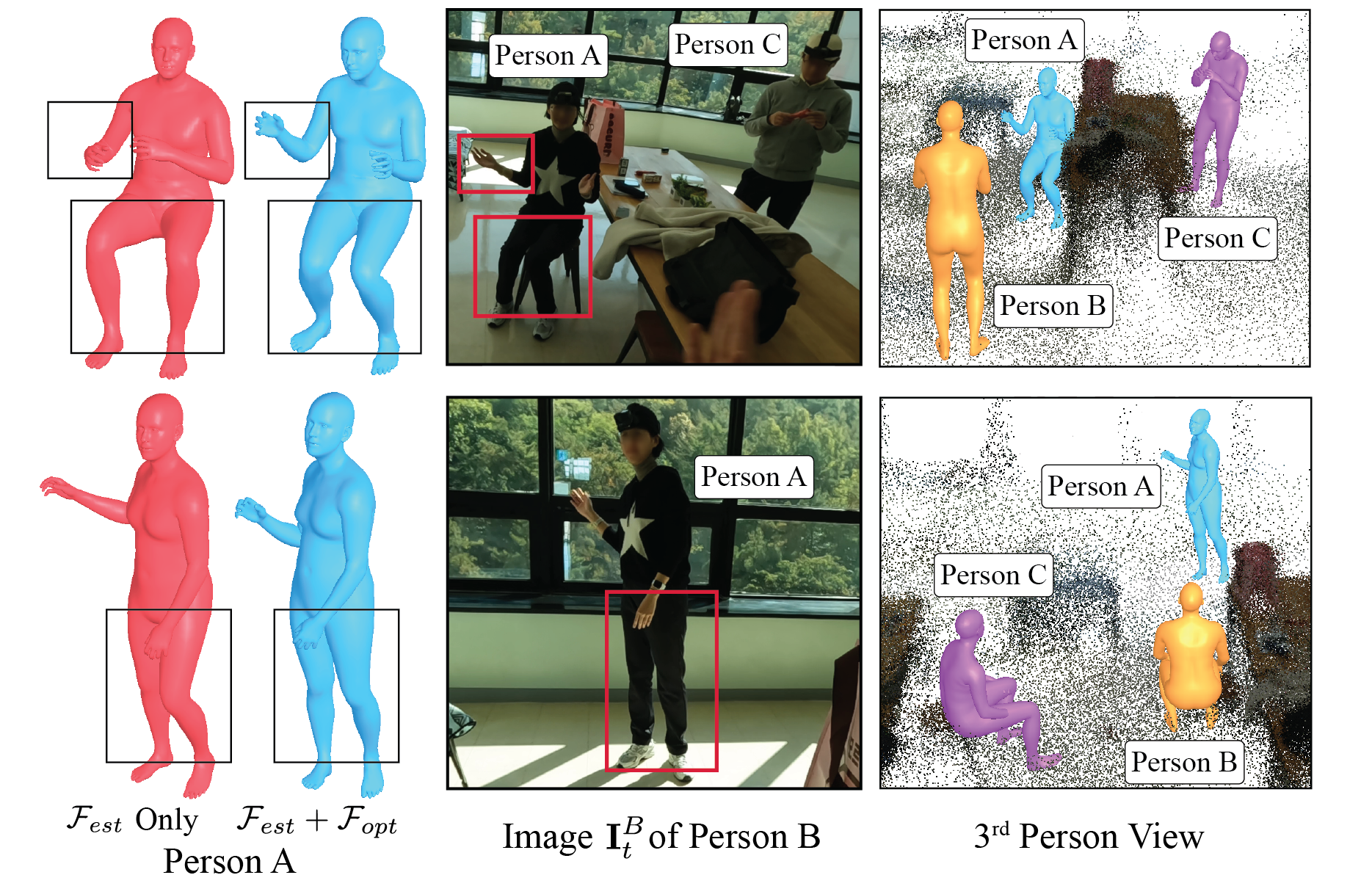}
    \caption{Results of $\mathcal{F}_{est}$ and $\mathcal{F}_{est} + \mathcal{F}_{opt}$ in multi-person scenarios. (a) Comparison with $\mathcal{F}_{est}$(red) and $\mathcal{F}_{est}+\mathcal{F}_{opt}$(blue) of person A with visual cue $\bm{\tau}_t^{\protect\overrightarrow{BA}}$ (Sec.~\ref{sec:motion_opt}). (b) Image $\mathbf{I}_t^{B}$ with person A. (c) 3rd person view of person A, B, and C.}
    \label{fig:multi_edit_abl}
    \vspace{-0.6cm}
\end{figure}

\noindent\textbf{Ablation on Motion Optimization.} We first demonstrate the contribution of $\mathcal{F}_{opt}$ in hand-based everyday interaction scenarios.
For quantitative comparison, we compare the initial motion output $\mathbf{M}_{reg}$ and optimized final output $\mathbf{M}$ to ground-truth.
As shown in Table~\ref{tab:edit_ablation}, right arm PE and both arm PE both decreased in the final motion output $\mathbf{M}$ compared to the initial regression-based output $\mathbf{M}_{reg}$. Example results are shown in Fig.~\ref{fig:edit_abl}. 
We also demonstrate motion optimization results from visual cues in multi-person scenarios. As seen in Fig.~\ref{fig:multi_edit_abl}, the egocentric image of person B serves as a ``third-person view" of person A. Using this information can resolve the fundamental ambiguities of the legs that stem from upper-body sensor setups.
Although the legs of $\mathcal{F}_{est}$ (red) in Fig.~\ref{fig:multi_edit_abl} (a) are plausible, $\mathcal{F}_{opt}$ (blue) further captures the slight nuances of leg movements by leveraging visual cues $\bm{\tau}_t^{\protect\overrightarrow{BA}}$.

\section{Discussion}
\label{sec:discussion}
We present a novel, easy-to-use, and affordable motion capture system with smartwatches on wrists and a head-mounted camera. We overcome the sparsity and ambiguity of sensor inputs with different modalities by integrating head poses in the motion estimation pipeline.
By tracking and updating floor levels to define head poses and incorporating into multi-stage Transformer-based estimation module our method can robustly capture motion in challenging locations with non-flat grounds.
We further present a motion optimization approach by using visual cues through the camera.
As multi-users can conveniently equip our setup, we also explore the scenarios when user signals are shared.

Although the off-the-shelf models (e.g., DROID-SLAM) in our method are robust and show reliable results in most cases, our method does not work in rare catastrophic failures of the off-the-shelf models.
Moreover, the networks in our module are trained with mean body shape, and head poses in real-world demonstrations are adjusted by scaling. 
Explicitly handling body shape variations in the module would be more promising which we leave as future work.
\vspace{-0.1in}

\paragraph{Acknowledgements.}
We thank Inhee Lee for supporting the system setup and thank Jiwon Song, Sumin Lee, and Hakjean Kim for assisting multi-person capture.
This work was supported by SNU Creative-Pioneering Researchers Program, NRF
grant funded by the Korea government (MSIT) (No.2022R1A2C2092724 and No. RS-2023-00218601), and IITP grant funded by the Korean government (MSIT) (No.2021-0-01343). H. Joo is the corresponding author.

{\small
\bibliographystyle{ieee_fullname}
\bibliography{egbib}
}

\clearpage 
\appendix
\section{Supplementary Video}
The supplementary video in the \href{https://jiyewise.github.io/projects/MocapEvery/}{project page} shows the real-world demonstrations of our motion capture system with two smartwatches and a head-mounted camera. The real-world demonstrations include various scenarios including expansive outdoor scenes, everyday motions and interactions (e.g., making coffee), motions with dynamic movements, and social interactions among multiple people. We also visually demonstrate floor level changes in areas with drastic floor level changes such as walking down the stairs. 

We demonstrate comparison with previous state-of-the-art methods that estimates motion from full-body IMU sensor setups~\cite{yi2022pip, jiang2022tip}. The video also includes several ablation studies of our method, showing the contribution of (1) floor level update module on non-flat scenes in module $\mathcal{F}_{est}$ (Sec.~\ref{sec:motion_est}) (2) motion optimization $\mathcal{F}_{opt}$ with visual cues $\phi_E$ and $\phi_T$ (Sec.~\ref{sec:motion_opt}).

\section{Additional Experiments}
\subsection{Camera-Derived Head Poses}

In the main paper, we have demonstrated the strength of our motion estimation module $\mathcal{F}_{est}$ against baselines, where we test with the head pose generated by available motion capture data given that paired egocentric videos are not available in the existing real IMU dataset.
As an extension of this experiment, we further perform a similar quantitative evaluation with egocentric videos using the dataset in EgoLocate~\cite{yi2023egolocate}, where the motions in the TotalCapture~\cite{trumble2017totalcapture} dataset are paired with egocentric videos taken from virtual head-mounted cameras in synthetic scenes.
This additional experiment provides further evidence of the strength of our method in a closer scenario with the practical setup, accounting for potential noise introduced during the head pose estimation process in monocular SLAM.
To demonstrate the reliability of the pre-processing stage of deriving head pose from egocentric video input, which happens prior to $\mathcal{F}_{est}$, we use the dataset in Egolocate~\cite{yi2023egolocate} where the motions in the TotalCapture~\cite{trumble2017totalcapture} dataset are paired with egocentric videos taken from virtual head-mounted cameras in synthetic scenes.

We randomly select 30 sequences in total (about 60000 frames), 15 for each scene. We follow the procedure in Sec.~\ref{sec:preprocess} to obtain head poses from the egocentric videos using DROID-SLAM~\cite{teed2021droid}. As the dataset does not undergo the alignment procedure in our system beforehand (Supp. Sec.~\ref{sec:setup}), the camera poses obtained from SLAM are aligned to world coordinates using head poses derived from the dataset. 

We first demonstrate the quality of the the camera-driven head poses from SLAM in Table~\ref{tab:head_cam}, where it shows minor differences from the head poses of the ground truth motion capture data; rotation error within 2-3 degrees and position error within 5 centimeters. 
This result can support the validity of our quantitative evaluations in our main paper.

In Table~\ref{tab:cam_dataset_head_comp}, we compare the motion estimation outputs with the 
camera-driven head poses, denoted as $\mathbf{H}^{C}+\mathcal{F}_{est}$, with the results with the methods with 6 IMU sensors, PIP~\cite{yi2022pip} and  TIP~\cite{jiang2022tip}, and our method $\mathcal{F}_{est}$ with head poses from the dataset.
Our results with camera-driven head pose from the raw egocentric data mostly outperform the competing methods which use the 6 IMU sensors, even though 
our setup with an egocentric camera and fewer sensors (2 IMUs on the wrists) is much more challenging compared to the previous methods.

We furthermore compare our results with EgoLocate~\cite{yi2023egolocate}, which leverages 6 IMU sensors on the full body for body pose estimation and an additional head-mounted camera for global translation localization.
Interestingly, despite the reduced number of sensors, our results outperforms EgoLocate.
Especially for root-related position error terms (MPJPE, Root PE) our method significantly outperforms.
Such improvement, despite of reduced number of sensors, initially stems from the accuracy of camera pose estimation from DROID-SLAM~\cite{teed2021droid} (Table~\ref{tab:head_cam}), and by directly incorporating the 6DoF head pose cues from the cameras to the motion estimation process, unlike EgoLocate that employs a decoupled approach between motion estimation (IMU) and localization (camera).

The discrepancy in the results of the root-included metrics between $\mathbf{H}^{C}+\mathcal{F}_{est}$ and $\mathcal{F}_{est}$ can be considered due to the camera position errors, as depicted in Table~\ref{tab:head_cam}.

\begin{table}
\footnotesize
    \centering
    \begin{tabular}{cccc}
    \toprule
        Scene & Rot. Error $(deg)$ & Pos. Error $(cm)$ & Frames \\
        \midrule
         \texttt{flood-ground} & 2.69 & 5.09 & 27026 \\
        \texttt{japan-office} & 2.12 & 4.73 & 32421 \\
    \bottomrule
    \end{tabular}
    \caption{Rotation and position error of camera-derived head poses compared to ground truth. (Frames are counted in 30 FPS)}
    \label{tab:head_cam}
\end{table}

\begin{table}
    \footnotesize
    \centering
    \begin{tabular}{c|c|c|c|c}
    \toprule
         Scene & Method & r.MPJPE & Root PE & MPJPE  \\
    \midrule  
     \texttt{flood-ground} & PIP~\cite{yi2022pip} & \underline{4.76} & 40.35 & 40.81  \\
          & TIP~\cite{jiang2022tip} & 5.13 & 42.06  & 41.53  \\
          & EgoLocate~\cite{yi2023egolocate} & 4.93 & 30.87  & 31.32  \\
   & $ \mathbf{H}^{C}+\mathcal{F}_{est}$ & \underline{4.76} & \underline{10.52} & \underline{11.44}  \\
          & $\mathcal{F}_{est}$ & \textbf{4.64} & \textbf{4.21} & \textbf{5.12}  \\
    \midrule  
     \texttt{japan-office} & PIP~\cite{yi2022pip} & \underline{4.34} & 27.20 & 27.76  \\
          & TIP~\cite{jiang2022tip} & 4.75 & 30.95  & 31.63  \\
          & EgoLocate~\cite{yi2023egolocate} & 4.43 & 23.35 & 23.72  \\
   & $\mathbf{H}^{C}+\mathcal{F}_{est}$ & 4.37 & \underline{9.76} & \underline{10.64}  \\
          & $\mathcal{F}_{est}$ & \textbf{4.13} & \textbf{3.75 } &\textbf{ 4.58}  \\
    \bottomrule
    \end{tabular}
    \caption{Comparison on motion estimation results with camera-derived head pose as input. $\mathbf{H}^{C}$ indicates head poses from camera trajectory of egocentric video $\mathbf{I}$. All metrics are in $cm$ scale. }
    \label{tab:cam_dataset_head_comp}
    \vspace{-0.2in}
\end{table}

\subsection{Ablation Studies on Motion Estimation} \label{sec:estimation_abl_baseline}

Although the motion estimation module $\mathcal{F}_{est}$ does not take the absolute position of wrists as input, our module $\mathcal{F}_{est}$ shows comparable or improved results compared to VR based baselines. (Table~\ref{tab:baseline_comparison_vr} of the main paper) To further clarify such result, we also conduct an ablation study on the motion estimation module to demonstrate the contribution of components in $\mathcal{F}_{est}$ for the performance. We first show the contribution of the multi-stage architecture (Sec.~\ref{sec:motion_est}) where the initial module, $\mathcal{F}^{ee}$, determines end-effector positions before estimating the whole body motion. Moreover, we highlight the contribution of data representation in input which combines normalized and global coordinates.
This combination of coordinates is advantageous as normalization standardizes actions and removes unwanted variations, while global information resolves ambiguities, like distinguishing between standing and sitting still. This dual-coordinate approach facilitates easier learning and enhances the network's robustness in making inferences. Results in Table~\ref{tab:abl_network} validate the contribution of each component.

\section{System Setup}\label{sec:setup}

\subsection{Sensors} 

\noindent\textbf{Readings IMU Sensor Signals from Smartwatches.} For demonstrations with smartwatches, we use Apple Watch SE2~\cite{applewatch} and the SensorLog app~\cite{sensorlog} to record and access IMU sensor data. From the recorded sensor signals, IMU rotations are obtained from ``motionQuaternion" and accelerations from ``motionAcceleration"
of the Apple Watch OS~\cite{applewatch_os_devkit}. The sensor data are recorded in 30 FPS.
Furthermore, we apply an average filter on the acceleration data (window length: 7). As previously demonstrated~\cite{jiang2022tip}, applying the average filter on both real and synthetic acceleration data makes the data sufficiently similar to each other. Note that our module is trained on synthetic IMU data, and we apply our trained model on real IMU data from smartwatches without additional fine-tuning, different from the previous methods~\cite{yi2021transpose, yi2022pip}.

\noindent\textbf{Head-Mounted Camera.} We use GoPro cameras for the head-mounted camera. To determine the timecode (in microseconds) when the camera shutter is pressed, we use the Open-GoPro Python SDK~\cite{opengopro} to activate the camera shutter.

\begin{table}
\footnotesize
\centering
\begin{tabular}{cc|ccc}
\toprule
data rep & multi-stage & MPJPE($\downarrow$) & r.MPJPE($\downarrow$) & MPJVE($\downarrow$) \\ \midrule
 & & 8.50 & 5.85 & 27.63 \\
 \(\checkmark\) & & 6.71 & 5.58 & 23.69 \\
\(\checkmark\) &  \(\checkmark\) & \textbf{5.20} & \textbf{4.95} & \textbf{17.00} \\
\bottomrule
\end{tabular}
\caption{Ablation studies on the components of $\mathcal{F}_{est}$. Settings are identical as in Tab.~\ref{tab:baseline_comparison_vr} (AMASS) in the main paper.}
\label{tab:abl_network}
\end{table}

\subsection{Calibration and Alignment} \label{sec:calib}
\begin{figure}
\centering
\includegraphics[width=0.9\linewidth]{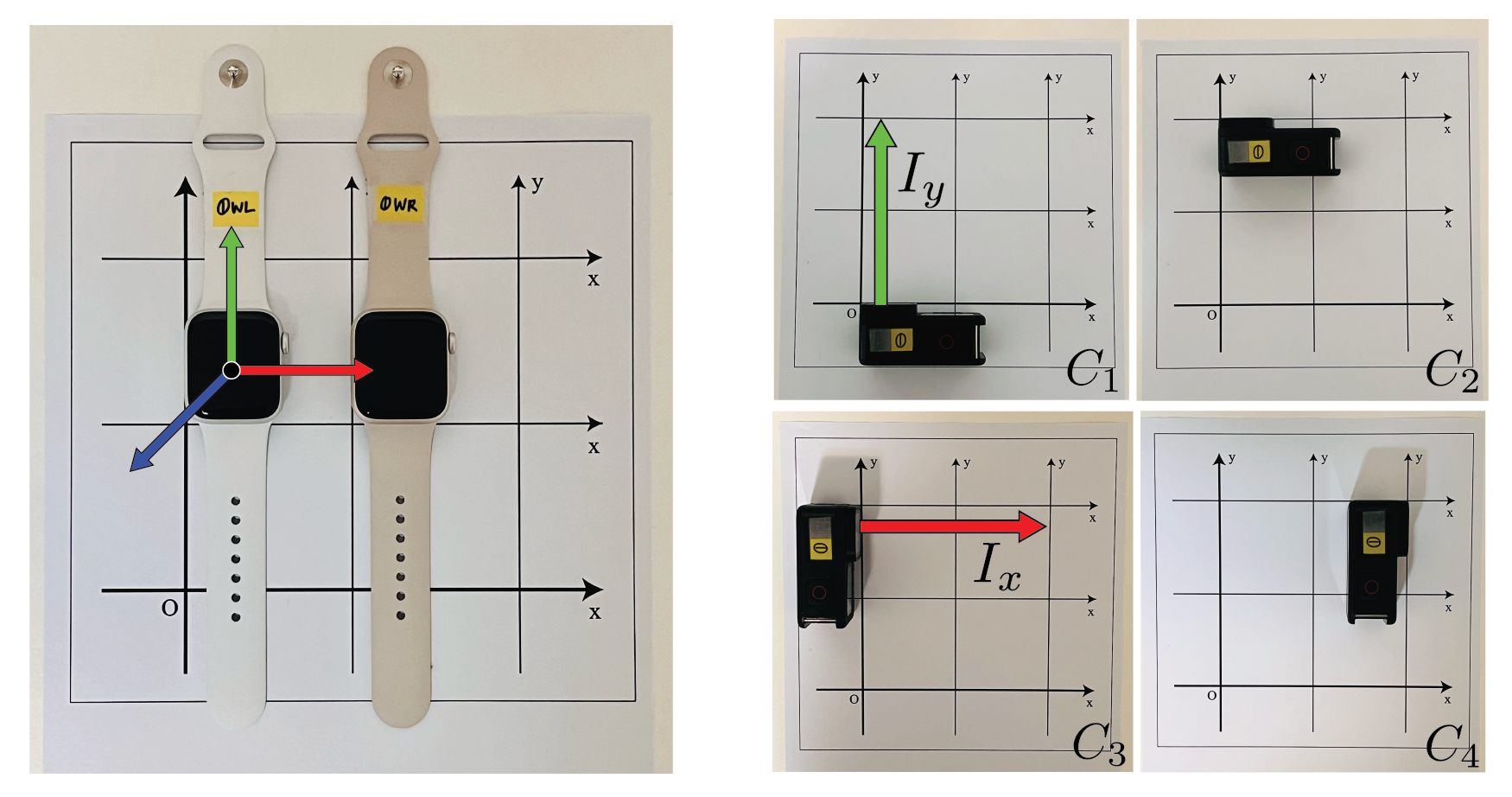}
    \caption{(a) Placing IMU sensors for IMU sensor calibration. The negative z-axis of the sensors are aligned to the gravity direction, and the y-axis is set as a ``facing direction". (b) Placing cameras on axes $I_x$ and $I_y$ for camera coordinate alignment.}
\label{fig:calib_system}
\end{figure}

\begin{figure}
\centering
\includegraphics[width=1.0\linewidth]{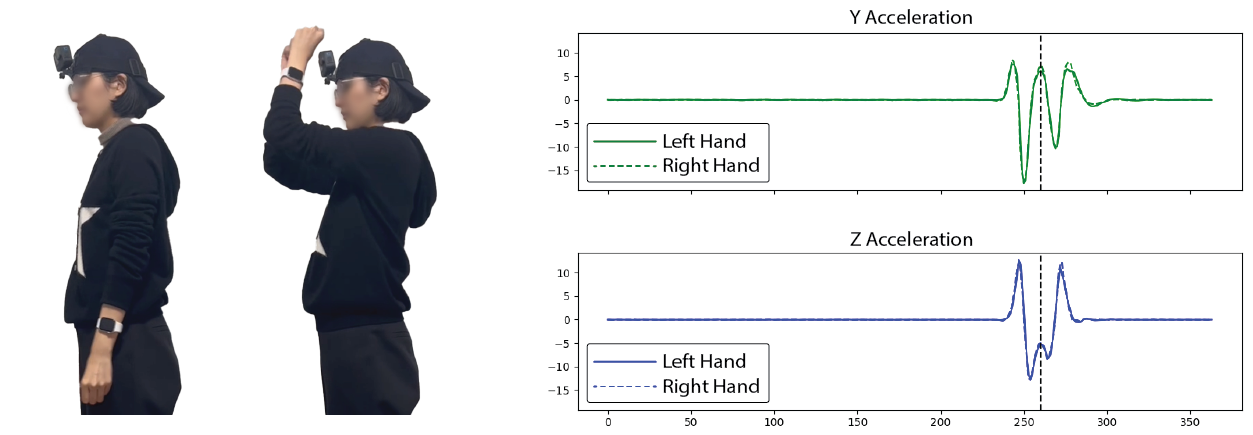}
    \caption{(a) Swinging arms for time synchronization. (b) IMU acceleration signals during sync. The vertical line indicates when the hand is up.}
\label{fig:timesync}
\vspace{-0.3cm}
\end{figure}

For coordinate alignment, the user has to follow 4 steps: (1) put the smartwatches (IMU sensors) on the calibration board for 3 seconds as in Fig.~\ref{fig:calib_system} (a); (2) wear the watches and stand in T-pose for 3 seconds; (3) put the camera on 4 positions on the calibration board as in Fig.~\ref{fig:calib_system} (b); (4) stand still for 10 seconds and swing arms up and down as in Fig.~\ref{fig:timesync} for time synchronization.

\noindent\textbf{IMU Sensor Calibration.}
In sensor data readings in real IMU sensors (e.g., smartwatches),
the IMU sensor data should first be calibrated to be coherent with the real-world coordinates where the gravity direction is in the negative z-axis. Here the y-axis is defined as the user's facing direction.
We use the notations similar to TIP~\cite{jiang2022tip}, which are as follows:
\begin{equation}
\begin{split}
    \mathbf{R}_{g}^{j} &= \mathbf{R}_{g}^{r}\mathbf{R}_{r}^{S}\mathbf{R}_{S}^{j}  \\
    \mathbf{a}_{g} &= \mathbf{R}_{g}^{r}\mathbf{R}_{r}^{S}\mathbf{a}_{r}
\end{split}
\end{equation}

\begin{figure}
\centering
\includegraphics[width=0.55\linewidth]{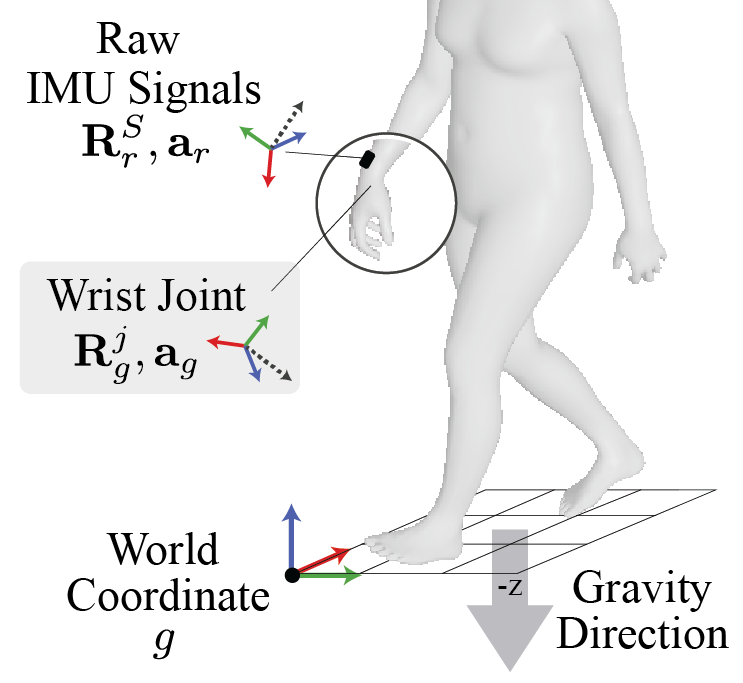}
    \caption{Visualization of the raw and calibrated IMU signals defined in reference coordinate $r$ and user-specified global coordinate $g$. }
\label{fig:imu_calib_vis}
\end{figure}

Let the reference coordinate where raw IMU sensors signals are defined as $r$, and the global coordinate coherent with real-world coordinates as $g$.
Finally, as the module $\mathcal{F}_{est}$ expects wrist joint rotations, the rotation value of IMU sensors should be calibrated into wrist joint rotations $\mathbf{R}_{g}^{j}$. Similarly, the raw acceleration data are defined in the reference coordinate of sensors and should be converted into coordinate $g$. Fig.~\ref{fig:imu_calib_vis} includes visualization of raw and calibrated signals.

$\mathbf{R}_{r}^{S}$ indicate rotations of raw IMU sensor data defined in the initial reference frame. At the first calibration stage, the user is asked to align the smartwatches to the specified coordinate $g$ (Fig.~\ref{fig:calib_system} (a)) so that 
in the moment $\mathbf{R}_{r}^{S} = \mathbf{R}_{r}^{g}$. From $\mathbf{R}_{r}^{g}$ we can derive $\mathbf{R}_{g}^{r}$ by applying $(\mathbf{R}_{r}^{g})^{-1}$.
Following previous methods~\cite{jiang2022tip, yi2021transpose}, the sensors are placed still for 3 seconds and the signal is averaged. Provided that the axes are aligned, it is not necessary for the two IMU sensors to be in the exact same position as IMU sensor signals are translation invariant.

The next calibration step is to obtain wrist rotations $\mathbf{R}_{g}^{j}$ from $\mathbf{R}_{g}^{S}$. This is done by applying $\mathbf{R}_{S}^{j}$, or $\mathbf{R}_{g}^{S} \mathbf{R}_{S}^{j}$. 
To obtain $\mathbf{R}_{S}^{j}$, the user wears smartwatches on both wrists and stands in T-pose, facing the y-axis in $g$. 
\begin{equation}
    \mathbf{R}_{S_T}^{j_T} = (\mathbf{R}_{r}^{S_T})^{-1}\mathbf{R}^{g}_{r}\mathbf{R}_{g}^{j_T}
\end{equation}
The subscript $T$ indicates the values obtained during T-pose calibration. Assuming that the IMU sensors are fixed in the wrist position during capture, we can assume that $\mathbf{R}_{S}^{j}$ is fixed, or $\mathbf{R}_{S_T}^{j_T} = \mathbf{R}_{S}^{j}$. 
As the joint rotations in T-pose, or $\mathbf{R}_{g}^{j_T}$, is known, $\mathbf{R}_{S_T}^{j_T}$ can be derived by obtaining raw sensor recordings during T-pose calibration $\mathbf{R}_{r}^{S_T}$. Similar to the first step, the user stands in T-pose for 3 seconds.

\begin{figure}
\centering
\includegraphics[width=0.9\linewidth]{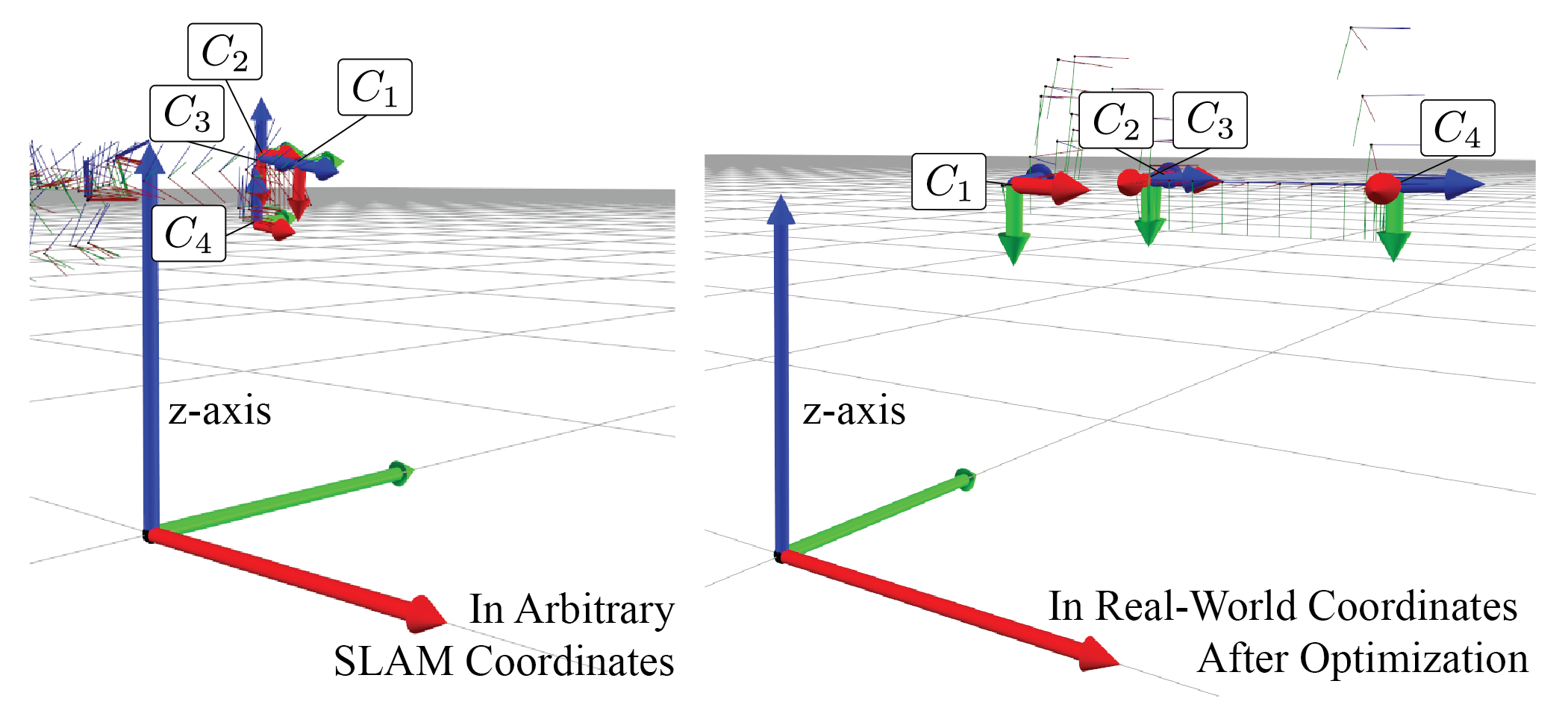}
    \caption{(a) Camera poses $C_i$ in arbitrary SLAM coordinates. (b) Camera poses $C_i$ in real-world coordinates after alignment optimization.}
\label{fig:calib_camera_opt_comp}
\end{figure}

Note that unlike previous methods~\cite{yi2021transpose, jiang2022tip} we do not consider acceleration bias (mostly, gravity) as the Apple Watch OS offers unbiased acceleration values.

\noindent\textbf{Camera Alignment to IMU Coordinates.}
For aligning camera coordinates to IMU coordinates, previous approaches~\cite{guzov2021hps, yi2023egolocate} leverage joint positions derived from IMU-based mocap modules for alignment. Different from such methods, our alignment method does not rely on IMU-derived body joint positions. 
At the start of recording, the user is instructed to put the camera in 4 positions
defined on the x and y axis, denoted as $I_x, I_y$,  of the global coordinate $g$ defined in IMU calibration. (Fig.~\ref{fig:calib_system} (b)).

The corresponding camera positions in arbitrary coordinates from SLAM are denoted as $C_i, i \in \{1, ..., 4\}$. 
Alignment is done by finding the transformation matrix $T_I^{c}$ which maps $\protect\overrightarrow{C_1, C_2}$ to y-axis $I_y$ and $\protect\overrightarrow{C_3, C_4}$ to x-axis $I_x$. 
As the real-world data could be noisy and a closed-form solution may not exist, we formulate an optimization problem to find the optimal transformation matrix.

The scale of arbitrary coordinates of SLAM is set by measuring the distances in the calibration board in Fig.~\ref{fig:calib_system}. The initial height $h_0$ is given by measuring the distance between the board and the floor the user stands on during the alignment procedure. The camera poses before and after alignment are in Fig.~\ref{fig:calib_camera_opt_comp}.

\noindent\textbf{Head Poses From Camera Trajectory.} 
The camera center $\mathbf{C}$ may not be necessarily the same as the head joint location, we compute the fixed transformation $T_{head}^{cam}$ to transform the camera pose into the head pose.
The translation component fo $T_{head}^{cam}$ is computed by approximating the camera location in a surface point of SMPL mesh. As the camera may not be on the surface of the human head as in Fig.~\ref{fig:timesync}, an offset is added to the surface point. Similar to the T-pose calibration step in IMU sensor calibration (Supp. Sec.~\ref{sec:calib}), the rotation component of $T_{head}^{cam}$, or $\mathbf{R}_{head}^{cam}$, is obtained by asking the user to stand still (as in T-pose) for 10 seconds. 

\noindent\textbf{Synchronization.}
The initial time synchronization between IMU sensors and the head-mounted camera is set by the timecodes generated from the sensors. (For the camera, the timecode is recorded when the shutter is activated) However, empirically we found there were minor errors in the timecode and therefore should be adjusted. For the adjustment, 
the user is asked to stand still and swing their arms up and down, as in Fig.~\ref{fig:timesync} (a). When the hands move up to the highest position close to the camera (Fig.~\ref{fig:timesync} (a)), the IMU sensor signals show a specific peak as in Fig.~\ref{fig:timesync} (b). The time synchronization is done by aligning the two. 
To adjust between the camera and the IMU sensors, the frame where the hands are closest to the camera is selected.
We empirically found out that swinging is more robust than clapping, as in some cases the force transmitted to the wrist sensor during clapping can cause sensor peaks to spike.

\subsection{Capture Setup for Ablations}
For quantitative evaluation of ablative baselines (Sec. 4.4) XSens MVN Link~\cite{xsens} was used to capture ground-truth data.
As an IMU-based method, however, XSens also suffers from root drift issues~\cite{guzov2021hps}. We ignore these factors during the evaluation by considering errors in XY aligned space (in evaluating floor update algorithm) or capturing motion with relatively small root translations (in evaluating $\mathcal{F}_{opt}$).

For the ablation of floor plane update, the height changes from 0m to $-$4.21m. For ablations on visual-cue based motion optimization $\mathcal{F}_{opt}$ in egocentric cases, we capture 3 scenarios (making coffee, using coffee machine, taking snacks off the shelf; total 2289 frames) and the right hand was detected as the visual cue $\phi_E$.

\section{Implementation Details}

\subsection{Synthesizing IMU Sensor Data} For synthesizing IMU sensor data from motion capture datasets we follow the protocol in~\cite{yi2021transpose, yi2022pip, jiang2022tip}. Rotations are obtained from joint rotations derived by solving forward kinematics. Accelerations are synthesized based on the following equation:
\begin{equation}
    \mathbf{a}_t^j = \frac{\mathbf{p}^j_{t - n} + \mathbf{p}^j_{t + n} - 2\mathbf{p}^j_{t}   }{(n\Delta t)^2} \quad j\in\{left, right\}
\end{equation}
We set $n = 4$, which is reported~\cite{yi2021transpose} to synthesize sensor signals closest to the real sensors.

\subsection{Floor Level Update}

\noindent\textbf{Implementation Details.} 
The threshold $\lambda$ for contact detection is set to 0.5.
For projecting $\mathbf{p}_{t_{m}}^{f}$ to the pointcloud $\mathbf{W}$, set points $\{\mathbf{w}\}$ whose distance to $\mathbf{p}_{t_{m}}^{f}$ are less than 0.15m are searched.
As the pointcloud $\mathbf{W}$ is not pre-scanned but is obtained from SLAM, the number of points may be sparse in floor regions. 
If the number $N$ of points in $\{\mathbf{w}\}$ is below 10, the height of $\mathbf{p}_{t_{m}}^{f}$ is set as $f_t$. 
To prevent undesired updates, the floor level is not updated when (1) the floor level change is less than 0.1m, (2) both feet have been in contact from $t-5$ to $t-1$, and are still in contact for $t$ to $t+5$, (3) the time interval between current and previous update is less than 25 frames.

\subsection{Motion Estimation} \label{sec:motion_est_supp}

\noindent\textbf{Models and Training.}
For the transformer encoders in both submodules $\mathcal{F}^{end}$ and $\mathcal{F}^{body}$ 
the number of heads is set to 10, and the number of layers to 4. For $\mathcal{F}^{end}$, the input is projected into embeddings of dimensions 1280. For $\mathcal{F}^{body}$, mid-representations $\{\mathbf{x}_\tau^{mid}\}$ and $\{\mathbf{x}_\tau\}$ are projected into embeddings of dimension 640 each and are concatenated and fed into the Transformer encoder. The features generated by the Transformer encoder are converted to the final output of each module via a 2-layer MLP, which converts the features into 256 dimensions in the first layer and to the dimensions of the final output in the second layer. 
The length $N$ of temporal sliding windows fed into the modules is set to 40. 
The two modules are trained end-to-end, with AdamW optimizer~\cite{loshchilov2017adamw}. The learning rate is automatically set with DAdaptation~\cite{defazio2023dadapt}. We use Nvidia RTX 3090Ti GPU for training. PyTorch~\cite{pytorch} and FairMotion library~\cite{gopinath2020fairmotion} were used for implementation.

\noindent\textbf{Datasets.} The datasets used for baseline comparison are stated in the main paper. For training the model used in real-world demonstrations in the supplementary video, we use a subset of AMASS (CMU~\cite{cmumocap}, HDM05~\cite{muller2007hdm05}, BMLMovi~\cite{ghorbani2020movi}, KIT~\cite{mandery2016kit}, HUMAN4D~\cite{chatzitofis2020human4d}) and a subset of Lafan1~\cite{harvey2020robust} dataset for training.

\subsection{Motion Optimization}

\noindent\textbf{Models and Training.}
The encoder and decoder of the autoencoder structure to build motion manifolds (Sec~\ref{sec:motion_opt}) which consist of 3 layers of 1D temporal-convolutions with a kernel width of 25 and stride 2. The channel dimension of each output feature is set to 256. The autoencoder is trained with the AdamW optimizer~\cite{loshchilov2017adamw} and the learning rate is automatically set with DAdaptation~\cite{defazio2023dadapt}.For optimizing the latent vector within the manifold, Adam optimizer~\cite{kingma2014adam} was used, and the learning rate was set to 0.0007 for single-person egocentric visual cues $\phi_E$, 0.001 for multi-person visual cues $\phi_T$. As in the motion estimation module, we use PyTorch~\cite{pytorch} and FairMotion library~\cite{gopinath2020fairmotion} for implementation. Nvidia RTX 3090Ti GPU was used for training and optimization. The datasets used for training the autoencoder are identical to the datasets used in training $\mathcal{F}_{est}$ (Supp. Sec.~\ref{sec:motion_est_supp}).

\noindent\textbf{Generating Visual Cues.} MediaPipe library~\cite{lugaresi2019mediapipe} was used for detecting 2D keypoints in single-person egocentric visual cues $\phi_E$. The missing keypoints are tracked using the optical flow of the detected keypoints. State-of-the-art monocular 3D pose estimation method 4DHumans~\cite{goel20234dhumans} was used for generating multi-person visual cues $\phi_T$.

\noindent\textbf{Reconstruction Loss.} The encoder $E$ and decoder $E^{-1}$ are trained based on reconstruction loss $\mathcal{L}_{recon} = ||\mathbf{X} - E^{-1}( E \left(\mathbf{X}) \right) ||$, where: 
\begin{equation}
\mathcal{L}_{recon} = \mathcal{L}_{contact} + \mathcal{L}_{root} + \mathcal{L}_{rot} + \mathcal{L}_{pos}.   
\end{equation}
$\mathcal{L}_{contact}$, $\mathcal{L}_{root}$, $\mathcal{L}_{rot}$, and $\mathcal{L}_{pos}$ are the L1 losses of foot contact labels,  root translation and rotation, joint rotations in 6D representations~\cite{zhou20196d}, and global joint positions obtained by forward kinematics operation.



\end{document}